\documentclass[journal,10pt]{IEEEtran}

\usepackage{latexsym}
\usepackage{balance}
\usepackage{epsfig}
\usepackage{cite}  
\usepackage{acronym}
\usepackage{amsfonts}
\usepackage{amssymb}
\usepackage{amsmath}
\usepackage{amsthm}
\usepackage{graphicx}                   
\usepackage{subfig} 
\usepackage{stfloats}
\usepackage{caption}
\usepackage{cases}                     
\usepackage{verbatim}                   
\usepackage{url}                      
\usepackage{algorithm}
\usepackage{algpseudocode}
\usepackage{color}
\usepackage{ifthen}
\newboolean{conclusions}
\setboolean{conclusions}{true}

\theoremstyle{plain}
\newtheorem*{Def}{Definition}

\usepackage{url}
\usepackage{xcolor}
\definecolor{newcolor}{rgb}{.8,.349,.1}

\usepackage{booktabs}

\usepackage{multirow}
\definecolor{orcidlogocol}{HTML}{A6CE39}
\usepackage{soul}
\algdef{SE}[DOWHILE]{Do}{doWhile}{\algorithmicdo}[1]{\algorithmicwhile\ #1}
\acrodef{ASP}{Associated Stochastic Problem}
\acrodef{EA}{Essential Attributes}
\acrodef{MI}{Mutual Information}
\acrodef{CE}{Cross-Entropy}
\acrodef{FE}{Feature Engineering}
\acrodef{MEC}{Mobile Edge Computing} 
\acrodef{mMTC}{massive Machine-Type Communications}
\acrodef{AV}{Autonomous Vehicle}
\acrodef{FFS}{Federated-Feature Selection}
\acrodef{NTN}{Non Terrestrial Network}
\acrodef{PLR}{Packet Loss Rate}
\acrodef{ES}{Edge Server}
\acrodef{FL}{Federated Learning}
\acrodef{PRR}{Packet Received Ratio}
\acrodef{BER}{Bit Error Rate}
\acrodef{RTT}{Round-Trip Time}
\acrodef{EDA}{Electrodermal Activity}
\acrodef{HR}{Heart Rate}
\acrodef{FS}{Feature Selection}
\acrodef{ADS}{Autonomous Driving System}
\acrodef{ECG}{Electrocardiography}
\acrodef{EEG}{Electroencephalography} 
\acrodef{EMG}{Electromyography}
\acrodef{BVP}{Blood Volume Pulse} 
\acrodef{ST}{Skin Temperature}
\acrodef{RSP}{Respiration} 
\acrodef{GD}{Gaze Detection}
\acrodef{HSM}{Human State Monitoring} 
\acrodef{ER}{Emotion Recognition}
\acrodef{SD}{Stress Detection}
\acrodef{WLD}{Workload Detection}
\acrodef{HAR}{Human Activity Recognition} 
\acrodef{HDC}{Heart Diseases Classification}
\acrodef{EOG}{Electrooculogram}
\acrodef{HAP}{High Altitude Platform}
\acrodef{IMU}{Inertial Measurement Unit}
\acrodef{AI}{Artificial Intelligence}
\acrodef{OEC}{Orbital Edge Computing}
\acrodef{TCP}{Transmission Control Protocol}
\acrodef{MCS}{Mobile Crowd-Sensing}
\acrodef{UDP}{User Datagram Protocol}
\acrodef{AIMD}{Additive Increase Multiplicative Decrease}
\acrodef{HOG}{Histogram of Oriented Gradient}
\acrodef{FAS}{Federated Aerospace System}
\acrodef{CPSoS}{Cyber Physical System of Systems}
\acrodef{CPS}{Cyber Physical System}
\acrodef{PEP}{Performance Enhancing Proxy}
\acrodef{CWND}{congestion window}
\acrodef{STP}{Satellite Transport Protocol}
\acrodef{HSMS}{Human State Monitoring System}
\acrodef{BDP}{Bandwidth-Delay Product}
\acrodef{QoS}{Quality of Service}
\acrodef{LoS}{Line of Sight}
\acrodef{NLoS}{Non Line of Sight}
\acrodef{BLoS}{Beyond Line of Sight}
\acrodef{QoE}{Quality of Experience}
\acrodef{ACM}{Adaptive Coding and Modulation}
\acrodef{DAMA}{Demand Assignment Multiple Access}
\acrodef{VANET}{Vehicular Ad-Hoc Network}
\acrodef{RA}{Random Access}
\acrodef{DA}{Dedicated Access}
\acrodef{OS}{Operating System}
\acrodef{CRA}{Contention Resolution ALOHA}
\acrodef{SA}{Slotted ALOHA}
\acrodef{DSA}{Diversity Slotted ALOHA}
\acrodef{OBU}{On-Board Unit}
\acrodef{CRDSA}{Contention Resolution Diversity Slotted ALOHA}
\acrodef{SIC}{Successive Interference Cancellation}
\acrodef{IC}{Interference Cancellation}
\acrodef{SINR}{Signal-to-Interference-plus-Noise Ratio}
\acrodef{ARQ}{Automatic Repeat reQuest}
\acrodef{SC-ARQ}{Selective-Coded ARQ}
\acrodef{SR-ARQ}{Selective-Repeat ARQ}
\acrodef{IRSA}{Irregular Repetition Slotted ALOHA}
\acrodef{CGC}{Complementary Ground Component}
\acrodef{RSU}{Road Side Unit}
\acrodef{ACK}{Acknowledgment}
\acrodef{S-NS3}{Satellite Network Simulator}
\acrodef{PMF}{Probability Mass Function}
\acrodef{NACK}{Negative Acknowledgment}
\acrodef{DVB-SH}{Digital Video Broadcasting - Satellite Services to Handhelds}
\acrodef{DVB-H}{Digital Video Broadcasting - Handheld}
\acrodef{DVB-RCS2}{Digital Video Broadcasting - Return Channel via Satellite}
\acrodef{SACK}{Selective Acknowledgment}
\acrodef{SNACK}{Selective Negative Acknowledgment}\acrodef{SNACK}{Selective Negative Acknowledgment}
\acrodef{SNIR}{Signal to Noise plus Interference Ratio}
\acrodef{SCPS-TP}{Space Communications Protocol Specifications - Transport Protocol}
\acrodef{CCSDS}{Consultative Committee for Space Data Systems}
\acrodef{ESA}{European Space Agency}
\acrodef{NASA}{National Aeronautics and Space Administration}
\acrodef{BSM}{Broadband Satellite Multimedia}
\acrodef{RLNC}{Random Linear Network Coding}
\acrodef{NC}{Network Coding}
\acrodef{FIFO}{First In, First Out}
\acrodef{FCFS}{First Come, First Served}
\acrodef{BLER}{Block Error Rate}
\acrodef{GEO}{Geosynchronous}
\acrodef{LEO}{Low-Earth Orbit}
\acrodef{FTP}{File Transfer Protocol}
\acrodef{CRC}{Cyclic Redundancy Check}
\acrodef{MAC}{Media Access Control}
\acrodef{HTTP}{Hypertext Transfer Protocol}
\acrodef{ISP}{Internet Service Provider}
\acrodef{MSS}{Maximum Segment Size}
\acrodef{BIC}{Binary Increase Congestion control}
\acrodef{AQM}{Active Queue Management}
\acrodef{XCP}{eXplicit Control Protocol}
\acrodef{ECN}{Explicit Congestion Notification}
\acrodef{CA}{Congestion Avoidance}
\acrodef{RED}{Random Early Detection}
\acrodef{TD}{Triple-Duplicate}
\acrodef{TO}{TimeOut}
\acrodef{MEO}{Medium-Earth Orbit}
\acrodef{NR}{new-radio}
\acrodef{IP}{Internet Protocol}
\acrodef{WMN}{Wireless Mesh Network}
\acrodef{ssthresh}{Slow-Start threshold}
\acrodef{MPE-IFEC}{Multi Protocol Encapsulation - Inter-burst Forward Error Correction}
\acrodef{FEC}{Forward Error Correction}
\acrodef{FSA}{Framed Slotted ALOHA}
\acrodef{D-FSA}{Diversity - Framed Slotted ALOHA}
\acrodef{CSA}{Coded Slotted ALOHA}
\acrodef{ML}{Machine Learning}
\acrodef{CRC}{Cyclic Redundancy Check}
\acrodef{P2P}{Peer-to-Peer}
\acrodef{EIRP}{Effective Isotropic Radiated Power}
\acrodef{FMT}{Fade Mitigation Technique}
\acrodef{SGD}{Smart Gateway Diversity}
\acrodef{NCC}{Network Control Centre}
\acrodef{ModCod}{Modulation and Coding}
\acrodef{FIFO}{First-In-First-Out}
\acrodef{WRR}{Weighted Round Robin}
\acrodef{WFQ}{Weighted Fair Queuing}
\acrodef{NS}{Network Simulator}
\acrodef{GSE}{Generic Stream Encapsulation}
\acrodef{PDF}{Probability Density Function}
\acrodef{CDF}{Cumulative Density Function}
\acrodef{AWGN}{Additive White Gaussian Noise}
\acrodef{CoV}{Coefficient of Variation}
\acrodef{MSC}{Message Sequence Chart}
\acrodef{ESA}{European Space Agency}
\acrodef{LIU}{Lebanese International University}
\acrodef{TUM}{Technical University of Munich}
\acrodef{MSCE-CS}{Master of Science in Communications Engineering - Communications Systems}
\acrodef{DLR}{German Aerospace Center}
\acrodef{NC-SGD}{Network Coding for SGD}
\acrodef{ATSP}{Advanced Transport Satellite Protocol}
\acrodef{PoI}{packet of interest}
\acrodef{STP}{Satellite Transport Protocol}
\acrodef{WMN}{Wireless Mesh Network}
\acrodef{SNR}{Signal-to-Noise Ratio}
\acrodef{SINR}{Signal-to-Interference-plus-Noise Ratio}
\acrodef{LMS}{Land Mobile Satellite}
\acrodef{LTE}{Long-Term Evolution}
\acrodef{M2M}{Machine-to-Machine}
\acrodef{IoT}{Internet of Things}
\acrodef{RA}{Random Access}
\acrodef{UAV}{Unmanned Aerial Vehicle}
\acrodef{UAS}{Unmanned Aerial System}
\acrodef{FANET}{Flying Ad-Hoc Network}
\acrodef{MANET}{Mobile Ad-Hoc Network}
\acrodef{VANET}{Vehicle Ad-Hoc Network}
\acrodef{C2}{Command and Control}
\acrodef{DTN}{Delay Tolerant Network}
\acrodef{COTS}{Commercial Off-the-Shelf}
\acrodef{IETF}{Internet Engineering Task Force}
\acrodef{CoAP}{Constrained Application Protocol}
\acrodef{MQTT}{Message Queue Telemetry Transport}
\acrodef{CoRE}{Constrained RESTful Environments}
\acrodef{ROLL}{Routing Over Low power and Lossy networks}
\acrodef{6Lo}{IPv6 over Networks of Resource-constrained Nodes}
\acrodef{URI}{Uniform Resource Identifier}
\acrodef{PUB/SUB}{Publish / Subscribe}
\acrodef{RCST}{Return Channel Satellite Terminal}
\acrodef{TDMA}{Time Division Multiple Access}
\acrodef{FDMA}{Frequency Division Multiple Access}
\acrodef{TCDMA}{Turbo Code Division Multiple Access}
\acrodef{PDMA}{Power Division Multiple Access}
\acrodef{WSN}{Wireless Sensor Network}
\acrodef{REST}{Representational State Transfer}
\acrodef{EDGE}{Enhanced Data rates for GSM Evolution}
\acrodef{UMTS}{Universal Mobile Telecommunications System}
\acrodef{LTE}{Long-Term Evolution}
\acrodef{E2E}{End-to-End}
\acrodef{3WHS}{Three-way Handshake}
\acrodef{SCADA}{Supervisory Control And Data Acquisition}
\acrodef{SOA}{Service-Oriented Architecture}
\acrodef{6LoWPAN}{IPv6 over Low power Wireless Personal Area Networks}
\acrodef{CoCoA}{CoAP Simple Congestion Control/Advanced}
\acrodef{RTO}{Retransmission TimeOut}
\acrodef{GPRS}{General Packet Radio Service}
\acrodef{TFRC}{TCP Friendly Rate Control}
\acrodef{DTMC}{Discrete Time Markov Chain}

\acrodefplural{RSU}[RSUs]{Road Side Units}
\acrodefplural{VANET}[VANETs]{Vehicular Ad-Hoc Networks}
\acrodefplural{OBU}[OBUs]{On-Board Units}
\acrodefplural{UAV}[UAVs]{Unmanned Aerial Vehicles}
\acrodefplural{FANET}[FANETs]{Flying Ad-Hoc Networks}
\acrodefplural{MANET}[MANETs]{Mobile Ad-Hoc Networks}
\acrodefplural{VANET}[VANETs]{Vehicle Ad-Hoc Networks}
\acrodefplural{UAS}[UASs]{Unmanned Aerial Systems}
\acrodefplural{RCST}[RCSTs]{Return Channel Satellite Terminal}
\acrodefplural{EA}[EAs]{Essential Attributes}

\begin{document}

\title{Feature Selection in Federated Remote Sensing for the Internet of Remote Things}
\title{Compressive and Federated Feature Learning in Distributed Remote Sensing}
\title{Federated Feature Selection for Data Compression at the Edge of the Internet}
\title{Federated Feature Selection for Cyber-Physical Systems of Systems}

\author{Pietro~Cassar\'a, Alberto Gotta, and Lorenzo Valerio
 \thanks{P. Cassar\'a and A. Gotta are with the National Research Council, Institute of Information Science and Technologies, Pisa,
PI, 56124 Italy e-mail: $\{$pietro.cassara@isti.cnr.it, alberto.gotta@isti.cnr.it$\}$}
\thanks{L. Valerio is with the National Research Council, Institute of Informatics and Telematics, Pisa,
PI, 56124 Italy e-mail: lorenzo.valerio@iit.cnr.it.}}

\markboth{IEEE Transactions on Vehicular Technology,~Vol.~, No.~, Month~Year}%
{Shell \MakeLowercase{\textit{et al.}}: Bare Demo of IEEEtran.cls for IEEE Journals}
%

\maketitle
\begin{abstract}
Autonomous vehicles (AVs) generate a massive amount of multi-modal data that once  collected and processed through Machine Learning algorithms, enable AI-based services at the Edge. In fact, not all these data contain valuable, and informative content but only a subset of the relative attributes should be exploited at the Edge. Therefore, enabling AVs to locally extract such a subset is of utmost importance to limit computation and communication workloads. Achieving a consistent subset of data in a distributed manner imposes the AVs to cooperate in finding an agreement on what attributes should be sent to the Edge.  
In this work, we address such a problem by proposing a federated feature selection algorithm where all the AVs collaborate to filter out, iteratively, the redundant or irrelevant attributes in a distributed manner, without any exchange of raw data. This solution builds on two components: a Mutual-Information-based feature selection algorithm run by the AVs and a novel aggregation function based on the Bayes theorem executed on the Edge.
Our federated feature selection algorithm provably converges to a solution in a finite number of steps.
Such an algorithm has been tested on two reference datasets: MAV with images and inertial measurements of a monitored vehicle, WESAD with a collection of samples from biophysical sensors to monitor a relative passenger.
The numerical results show that the fleet finds a consensus with both the datasets on the minimum achievable subset of features, i.e., 24 out of 2166 (99\%) in MAV and 4 out of 8 (50\%) in WESAD, preserving the informative content of data.
\end{abstract}


\IEEEpeerreviewmaketitle

\section{Introduction}

\IEEEPARstart{A}{utomation} enables a \ac{CPSoS} to run with a minimum human assistance and evolves into autonomy when the human is taken out of the sensing, decision, and actuation loop. 
Automation can be used to operate a \ac{CPSoS} comprising complex, dynamic, virtual and physical resources, such as telecommunication networks, computing units, software, sensors, and machines \cite{bondavalli2016cyber}. Humans can interact with an autonomous system either as passive end-users (such as passengers in autonomous transportation system) or rather as active co-operators in a mutual empowerment relationship towards a shared goal. Such cooperative, connected, and autonomous systems have the potential to be a game-changer in multiple domains if they will be capable of positively exploiting such an inescapable human factor. 
The increasing development of semi-\acp{ADS} poses the challenge of taking the end-user, in the middle of the evolution process toward fully \acp{ADS}. 
Aside from vehicle control, a \ac{CPSoS} needs to monitor the comfort/discomfort of the passenger, as well, to improve its well-being and to acknowledges the degree of safety and satisfaction perceived about the \ac{ADS}.
\ac{AI} is a fundamental technology for deploying the future \ac{CPSoS} for \acp{ADS} \cite{bacciu2021teaching}. The stringent computational and memory requirements for \ac{ML} algorithms will impose a significant rethinking of the underlying computing and communication system and will have to fit the constraints of the onboard units.
Information extraction should follow as much as possible optimal criteria, cooperating with the inherently distributed nature of the automotive scenario.

Moreover, local processing of information can also be an advantage in specific scenarios with intermittent connectivity or when data privacy is a key issue \cite{siegel2017survey}. 
Hence, reducing the transfer time needed of either raw data or the relative features is of the utmost importance in determining the performance of computation offloading. Intuitively, traditional data compression techniques \cite{qingqing2019visual} could reduce such a delay component, but will also degrade the relative classification performance \cite{xie2019source}, prolonging the training phases as well as degrading the inference performance.

Conversely, when information extraction algorithms produce massive streams of features, selecting the most relevant ones to feed a \ac{ML} model becomes very convenient, both in terms of compression and accuracy preservation.
Such an operation is known as \ac{FS} \cite{Saeys2007} and allows for achieving simpler and, therefore, more efficient \ac{ML}-based models \cite{egea18}. 

This work focuses on feature selection efficiency within a fleet of \acp{AV}, which collect, through their sensors, multi-modal raw measurements to be pre-processed and delivered to feed a remote edge server for inference tasks. Such a procedure can introduce information redundancy, which leads to a waste of computing and communication resources. 
The \ac{AV} ensemble aims at limiting the transmission to the top relevant features only.
However, just a subset of the top ones can be extracted from each local data collection in a distributed manner w.r.t. the whole set achievable from the union of the local datasets but in a centralized manner. In fact, the former case may lead to an inconsistent model \textit{w.r.t.} to the latter. 
Therefore, the \acp{AV}, shall participate to a collaborative \ac{FS} process, in order to exploit the whole information in a federated manner. 
We tackle this problem, by proposing, for the first time, a \ac{FFS} algorithm, exploiting a distributed computing paradigm applied to \acp{AV}.
In \ac{FFS} all \acp{AV} collaborate to come up with the minimal set of features selected from their local datasets.

\begin{figure}
    \centering
    \includegraphics[width=\columnwidth,trim={0 0 0 0},clip]{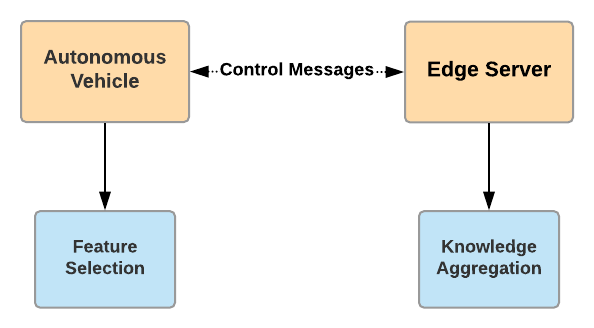}
\caption{Feature selection and aggregation components of the proposed \ac{FFS} system.}
\label{fig:FFS}
\end{figure}
The proposed \ac{FFS} system is made up of two components provided in Figure \ref{fig:FFS}:
\begin{itemize}
    \item a local \ac{FS} process runs on each \ac{AV} and aims at generating a local distribution probability that ranks the information associated to a given feature, according to the \ac{MI} metric \cite{brown09,Nguyen2014}, which is solved in this work by using the \ac{CE}, a suited method \cite{rubinstein04} to run in a distributed manner. 
    \item An aggregation algorithm executed on the \ac{ES} that combines the local estimates received from the \acp{AV}. The aggregation algorithm is based on a Bayesian approach to merge the local information into a global one. 
\end{itemize}

The messages delivered by \acp{AV} contain probability vectors where each element is the probability to select that feature. The ES returns the "federated" probability vector which is derived by the aggregation of the vectors received by the \acp{AV}, as detailed in the following, to replace each of the local ones.
Note that, the proposed approach does not need to share any local raw data but only the estimates of the local most informative features. Moreover, it guarantees that all the \acp{AV} reach a consensus on the subset of the most informative features, after a finite number of communication rounds, i.e., the messages exchanged between the \acp{AV} and the \ac{ES}. 

As we show in the paper, the proposed algorithm \textit{(i)} significantly limits the control messages exchanged during the \ac{FFS} process and \textit{(ii)} provably let the \acp{AV} converge to a subset of top features, which effectively reduce the information stored and transmitted by the \acp{AV}. Specifically, numerical results show that, on reference benchmarks, our solution limits data processing and transmission, by removing up the to 99\% of redundant features from the selected datasets, without loss of accuracy on the learning model.

Summarising, the novel contributions of this paper are: 
\begin{itemize}
    \item A novel \ac{FFS} algorithm based on the \ac{MI} \ac{CE} (client-side) on the \ac{AV} and a Bayesian aggregation approach on the \ac{ES}.
    \item The theoretical proof that such an algorithm converges to a stable solution in a fixed number of iterations. 
    \item An extensive numerical evaluation tested on two real-world datasets that shows the efficiency of our solution.
\end{itemize}





The paper is organized as follows: related works are presented in Section~\ref{sec:rel_works}; the reference  scenario  and the  system  assumptions are presented in Section~\ref{sec:problem_statement}; the theoretical background underlying the proposed feature selection approach is presented in Section~ \ref{sec:FE_theory_MAIN}; the federated version of the feature selection algorithm in presented in Section~\ref{sec:fed_approach}; Section~\ref{sec:num_results} presents the experimental results of a study case with two real world datasets, belonging to different application domains; conclusions in Section~\ref{sec:conclusion}.


\section{Related Works}
\label{sec:rel_works}


\subsection{Feature Selection}

Many \ac{FS} procedures have been proposed in the literature. In \cite{Jovic2015,Venkatesh2019,Saeys2007} authors provide a comprehensive overview of the existing methods. Additionally, they consider the most important application domains and review comparative studies on feature selection therein, in order to investigate, which methods outperform for specific tasks. 
Authors highlight that \ac{FS} is based on the identification of the relevance and redundancy provided by the features with respect to a class attribute function. The main approaches of \ac{FS} fall into three categories: filtering, wrapping, and embedded methods. This categorisation is based on the interaction between the selected features and the learning model adopted to take a decision. The output of the wrapping and embedded methods is tightly connected to the learning model that uses the selection. Therefore, with these methods \ac{FS} and model training cannot be uncoupled.
Conversely, filtering methods are suitable for being used regardless the presence of a learning model to train.  

As shown in \cite{Jovic2015,Venkatesh2019,Saeys2007}, most of the well-known filtering algorithms use information-based metrics for \ac{FS}, and can deal with samples of variable lengths, as presented in \cite{Chandrashekar2014,Miao2016}. A suitable information-based metric for the \ac{FS} is the \ac{MI}. \ac{MI} has gained increasing popularity in data mining, for its ease to use, effectiveness, and strong theoretical foundation.
mRMR \cite{mrmr} and HJMI \cite{hjmi} are some of the most used methods that exploit \ac{MI}. These approaches rank the features according to the maximization of the \ac{MI} and let the user to select a desired subset $k$. Differently, the proposed algorithm automatically select a minimal subset of relevant features, also capturing the mutual dependencies. Note that the formulation of the underlying optimization problem is NP-Hard\cite{brown09,Nguyen2014}, i.e., \ac{MI}-based feature selection problem involves the integer programming or, in some cases, the quadratic integer programming. In \cite{petroccia2019,anastasi2014,guarino2015} authors show how to adopt the \ac{CE} approach to address such native computational complex problems, for different application scenarios.
Beyond \ac{MI}, other filtering methods can use different metrics, such as in \cite{heiberger2009one} where the authors evaluate
the variance of all the features to measure the impact that each of them has on the learning process.
This method relies on the concept that the features with zero variance add no information, by considering the relation between the target variable and feature vectors. 

To the best of our knowledge, all these algorithms are designed for being executed in a centralised setting, i.e., under the assumption that the whole dataset is available to the learning agent. 

\subsection{Distributed Learning}
Distributed learning is considered from several perspectives in the literature.
A very consistent body of work deals with distributed learning based on the \ac{FL} framework. \ac{FL} is a distributed learning framework initially proposed by Google, where a large number of mobile or edge devices participate in a collective and distributed training of a shared model. \cite{Konecny2015,McMahan2016c}. 
\ac{FL} is an iterative procedure spanning over several communication rounds until the convergence is reached. Based on this paradigm, several modifications have been proposed concerning (i) new distributed optimisation algorithms~\cite{Wang2018,Amiri2019, Karimireddy2019,Mohri2019}, and (ii) privacy-preserving methods for \ac{FL}~\cite{Mao2018,MOTHUKURI2021619}. Alternatively, other approaches do not rely on a centralised coordinating server. In~\cite{Valerio:2016aa,Valerio:2017ab}, authors propose a distributed and decentralised learning approach based on Hypothesis Transfer Learning. Similarly to the \ac{FL} framework, authors assume that several devices hold a portion of a dataset to be analysed by some distributed machine learning algorithms. The aim of \cite{Valerio:2016aa,Valerio:2017ab} is to provide a learning procedure able to train, in a decentralised way, an accurate model while limiting the network traffic generated by the learning process.
The vast majority of the distributed learning solutions, presented in the literature, focus on the model's training, giving the feature engineering phase for granted. Until now, the idea of performing \ac{FS}, directly, on edge devices remains unexplored.

In the literature only few approaches cope with \ac{FS} in distributed settings. 
In \cite{ye2019}, authors present a distributed algorithm for \ac{FS} based on the Intermediate Representation, which aims at preserving the privacy of data, allowing the node to exchange each other the data they hold. Therefore, in this method \ac{FS} is performed under the assumption that all data are available to the \ac{FS} algorithm.
Moreover, the method presented by the author depends from the specific learning model that uses the selected features. 

In \cite{Banerjee2021}, the authors propose an information-theoretic \ac{FFS} approach called Fed-FiS.
Fed-FiS estimates feature-feature mutual information and feature-class mutual information to generate a local feature subset in each user device. Then a central server ranks each feature and generates a global dominant feature subset using a classification approach.
This approach has some commonalities with ours, such as the adopted metric (\ac{MI}) and the federated settings. However, differently from \cite{Banerjee2021} \textit{(i)} we provide directly the minimum set of relevant features instead of a ranking, \textit{(ii)} we propose an aggregation based on Bayes' theorem that does not rely on any Machine Learning scheme to finalise the selection (i.e., no regression or classification methods are adopted in our solution), resulting in a computationally more suitable approach for vehicular scenarios. 

In light of this and to the best of our knowledge, this is the first paper that proposes a federated mechanism of feature selection explicitly designed to meet the requirements of the \ac{CPSoS} context.

\section{System assumptions}
\label{sec:problem_statement}
\begin{figure}[tb]
\centering
    \includegraphics[width=1\columnwidth,trim={80 30 80 0},clip]{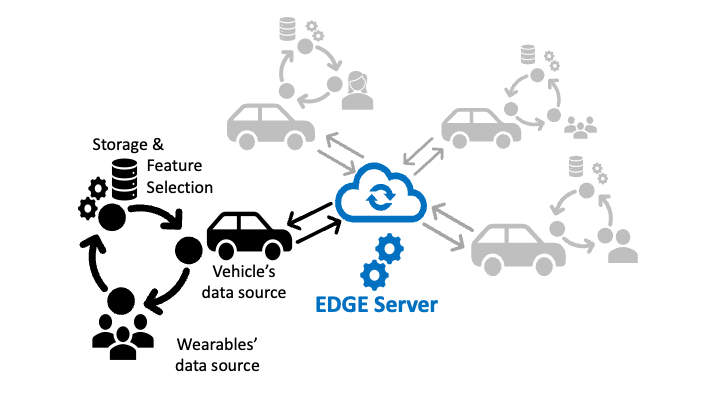}
\caption{System architecture. Data sources characterize two different \acp{CPS}: the former that monitors the user through wearable sensors, the latter relative to the \ac{ADS}.}
\label{fig:sys_arch}
\end{figure}

In this section, we describe the reference scenario and the system assumptions considered in this paper. As shown in Figure \ref{fig:sys_arch}, we consider a set of \acp{AV}, implementing an \ac{ADS} each, collecting data generated by the sensors integrated in a \ac{CPSoS} and that collaborates with the others \acp{ADS} to learn a minimal, and most informative set of features from their local datasets. To this end, the \acp{AV} execute an in-network data filtering process through our \ac{FFS} approach
to reach a consensus in identifying the most informative feature subset. Finally, the globally shared feature set is used like a compression scheme before transmitting it to an \ac{ES}. Note that, in this system the \acp{AV} are only responsible for finding the best compression scheme applicable to the their local data in a collaborative way, based only on the control information they exchange with the \ac{ES}. Moreover, the \ac{ES} has a three-fold role: i) it acts as central coordinating entity in the \ac{FFS} process whose purpose is to aggregate the partial control information sent by the \acp{AV}; ii) it acts as final collector for the compressed data, once the \ac{FFS} is completed and, iii) runs the \ac{AI} services to extract knowledge from data but that is used only for performance evaluation in this paper. 
We target two different user cases to validate the performance of the proposed \ac{FFS} method. The former refers to the localization of an \ac{AV} in the environment based on images and inertial measurements, and the latter regards the physiological-state monitoring of a passenger in the automotive domain. 
We define two different sub-systems part of the same \ac{CPSoS}: the \ac{ADS} of above, and an \ac{HSMS} to learn the feeling perceived from a passenger relatively to the \ac{ADS} driving style. Therefore, we assume each \ac{AV} to be equipped with a camera to capture images from the surrounding environment aside some inertial sensors for the former learning task, and a set of body sensors, such as, \ac{ECG}, \ac{EDA}, \ac{EMG}, and \ac{RSP} for the latter. 

Each \ac{AV} is able to locally synchronize the multi-sensory data such that, for each image, it is possible to associate the corresponding inertial measurements leading to an \emph{enhanced Raw Input Datum} (eRID). 
Note that for the scope of this paper it is not important the specific semantic of the labelling, but it is enough to assume a labelling process on the collected data. The \acp{AV} are also equipped with a relatively small edge computing unit (e.g., a RaspeberryPi or, at most, an Nvidia Jetson Nano) able to cache data and execute the \ac{FS} task, before transmitting the features. Additionally, the \acp{AV} are endowed with a radio communication interface to communicate toward the \ac{ES}. It must be noted that the task is not collecting images of the environment, or physiological parameters of the user but, conversely, retrieving the information associated to those images or to those physiological sensors, e,g., the position of the \ac{AV} with respect to the surrounding or the user mood. In particular, the latter is labelled according to the classification scale provided by questionnaires like PANAS, SSSQ or SAM \cite{Schmidt2018}, which associates numerical labels to the physiological states.

\section{Feature selection}
\label{sec:FE_theory_MAIN}
In this section, we provide the theoretical background of the \ac{MI}-based \ac{FS} algorithm and the relative implementation based on the \ac{CE} method.

\subsection{Background feature selection based on Mutual Information}
\label{sec:FE_theory}
To make the paper self-contained, we report in this Section the necessary theoretical background needed to get an intuition about the internal details of the CE-based \ac{FS} method presented in Section \ref{sec:ce-fs}. 

First, let us define the \ac{FS} problem as follows:
\begin{Def}[\ac{FS} Problem]
\label{def:fs}
Given the input data matrix $\mathbf{X}$ composed by $n$ samples of $m$ features ($\mathbf{X} \in \mathbb{R}^{n \times m}$), and the target attributes' (or labels) vector $\mathbf{y} \in \mathbb{R}^{n}$, the \ac{FS} problem is to find a $k$-dimensional subset $\mathbf{U} \subseteq \mathbf{X}$ with $k \leq m$, by which we can characterize $\mathbf{y}$.
\end{Def}
The method we adopt in the paper performs the \ac{FS} measuring, through the Mutual Information metric, the amount of information that a subset of features (or attributes) $\mathbf{U}$ expresses with respect to a specific target label $\mathbf{y}$.

Formally, the \ac{MI} between random variables can be defined as \cite{eliece77, cover91}: 

\begin{equation} 
  \label{MI_3}
  \mathbf{I}(\mathbf{U};\mathbf{y})=\mathbf{H}(\mathbf{y})-\mathbf{H}(\mathbf{y}|\mathbf{U}),
\end{equation}

where $\mathbf{U}=\{\mathbf{x}_1 \cdots \mathbf{x}_k \; | \; k\leq m\} \subseteq \mathbf{X}  $, and $\mathbf{H}(\mathbf{y}|\mathbf{U})$ is the conditional entropy which measures the amount of information needed to describe $\mathbf{y}$, conditioned by the information carried by $\mathbf{U}$. Hence, $ \mathbf{I} (\mathbf{U};\mathbf{y})$ represents the dependence between $\mathbf{U}$ and $\mathbf{y}$, i.e., the greater the value of $\mathbf{I}$, the greater the information carried by $\mathbf{U}$ on $\mathbf{y}$. We recall that the \ac{MI} between two random variables $\mathbf{A}$ and $\mathbf{B}$ is strictly related to the entropy $\mathbf{H}(\cdot)$, which defines the amount of information held by the variables, i.e., the entropy of a random variable $\mathbf{A}$ (i.e., $\mathbf{H}(\mathbf{A})$) and its probability are inversely proportional: the greater the entropy of a random variable $\mathbf{A}$, the greater its unpredictability and vice-versa. Hence, we can assert that the entropy measures the diversity of $\mathbf{A}$ in terms of the uncertainty of its outcomes. 

In MI-based \ac{FS} the features to be selected are those that maximise Equation (\ref{MI_3}). These features are typically referred as \ac{EA}. By solving the following optimization problem we would obtain the optimal global solution to the \ac{FS} problem defined in \ref{def:fs}:

\begin{gather} 
\label{Native Optimization Porblem}
\displaystyle \mbox{arg}\max_{\mathbf{U}} \mathbf{I}(\mathbf{U};\mathbf{y}) \\
\nonumber
\mathbf{U}=\{\mathbf{x}_1 \cdots \mathbf{x}_k \; | \; k\leq m\} \subseteq \mathbf{X}
\end{gather}%

Note that the problem (\ref{Native Optimization Porblem}) belongs to the class of Integer Programming (IP) optimization problems and finding its optimal solution is NP-hard \cite{Chaovalitwongse2009}, i.e., the optimal solution $\mathbf{U}$ would be found among all combinations of feature indices of the native set $\mathbf{X}$.

\begin{figure}[ht]
\begin{center}
  \includegraphics[width=1\columnwidth,trim={0 0 0 0},clip]{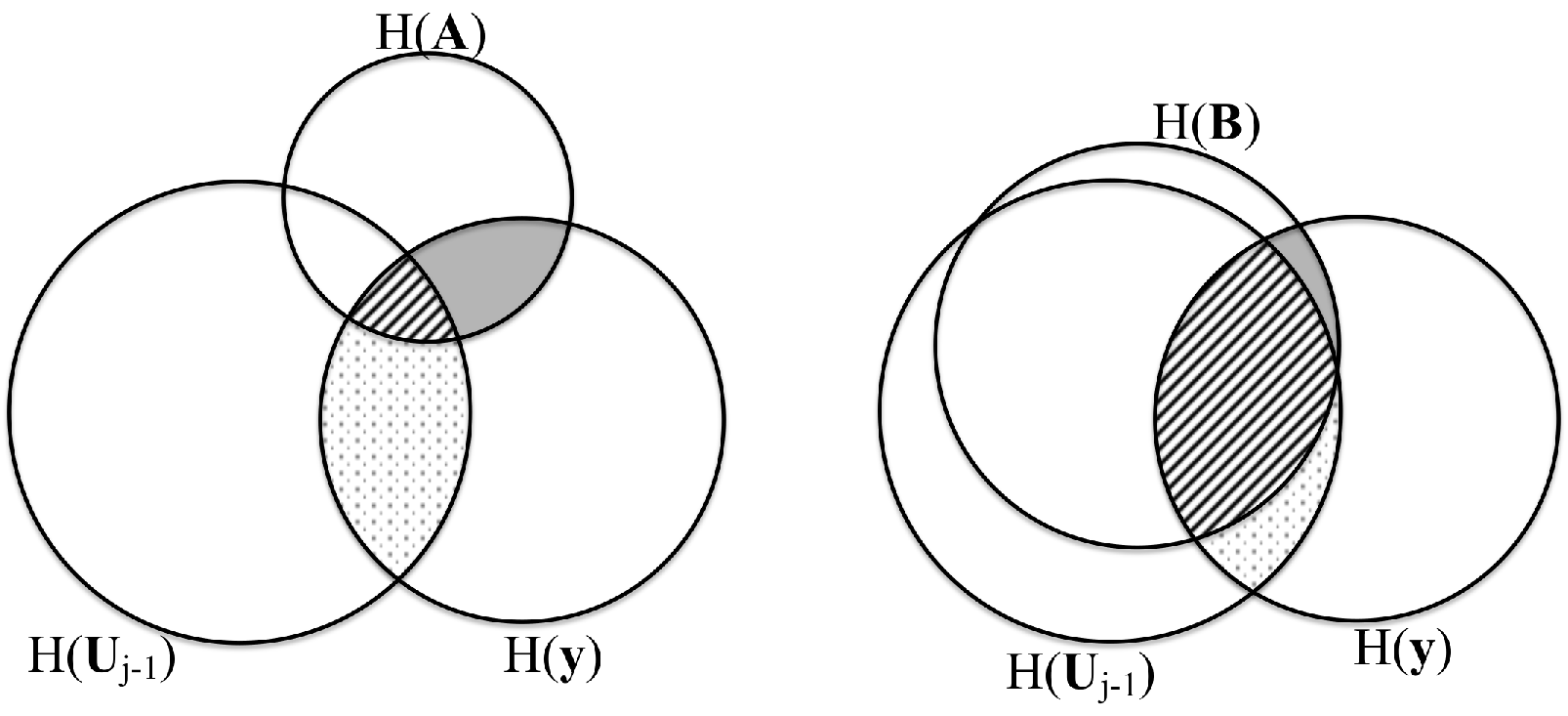}
  \caption{Example of the relationship between Mutual Information and Entropy} \label{figure_example} 
 \end{center}
 \end{figure}

The problem (\ref{Native Optimization Porblem}) becomes computationally tractable if approached through an iterative algorithm which selects and adds to the subset $\mathbf{U}$ one feature at a time. Therefore, instead of solving \ref{Native Optimization Porblem}, we address the problem defined in (\ref{Incremental Optimization Porblem}):
\begin{gather} 
\label{Incremental Optimization Porblem}
\displaystyle \mbox{arg}\max_{\mathbf{x}_j \in \mathbf{X}\setminus\mathbf{U}} \mathbf{I}(\mathbf{x}_j;\mathbf{y}|\mathbf{U}), \\
\nonumber
\mathbf{U}=\{\mathbf{x}_1 \cdots \mathbf{x}_{k-1} \; | \; k\leq m\} \subseteq \mathbf{X}.
\end{gather}
For the sake of clarity, we provide an intuitive example based on the relation between \ac{MI} and the entropy. Considering Figure \ref{figure_example}, the circles are the entropy of the random variables $\mathbf{A},\textbf{B},\mathbf{U},\mathbf{y}$, and the grey regions are the information carried by the variable $\mathbf{A}$ (or $\mathbf{B}$) on $\mathbf{y}$. The dashed area shows the information redundancy of the variable  $\mathbf{A}$ (or $\mathbf{B}$)  given the already selected variables in $\mathbf{U}_{j-1}$. In this example, the variable $\mathbf{A}$ should be added to the set $\mathbf{U}$ since it is more informative  than $\mathbf{B}$ on  $\mathbf{y}$ , i.e., its grey area is larger than $\mathbf{B}$'s, and it is less redundant than $\mathbf{B}$ w.r.t. to $\mathbf{U}_{j-1}$.

The main drawback of this approach is that it might end up with a sub-optimal solution because, by selecting the features one by one, the algorithm makes the implicit assumption that they are independent, which might not hold true. Theoretical foundations for the incremental version of the \ac{FS} algorithms has been proven by the authors in \cite{cover91, eliece77}. It is worth mentioning that a connected issue with problem (\ref{Incremental Optimization Porblem}) regards the efficient evaluation of the \ac{MI}, which might become prohibitive even for datasets with a small number of samples. We overcome this problem by adopting the MIToolbox \cite{Brown2012}, a state-of-the-art tool for numerical optimization. 


\subsection{CE-based feature selection algorithm}
\label{sec:ce-fs}
In this section, we describe the CE-based algorithm that finds, in a finite number of steps, a solution that well approximates the one found by solving problem (\ref{Native Optimization Porblem}), while making negligible the assumption of independence among features introduced in problem (\ref{Incremental Optimization Porblem}). In other words, with CE-based \ac{FS}, instead of selecting one \ac{EA} at a time, we select a set of \acp{EA} \emph{jointly}. 

The CE-based algorithm is based on the following intuition: if the set $\mathbf{U}$ contains only \acp{EA}, then $\mathbf{I}(\mathbf{U}; \mathbf{y}) \rightarrow \mathbf{H}(\mathbf{y} )$, which implies that $\mathbf{H}( \mathbf{y}| \mathbf{U}) \rightarrow 0$~\cite{cover91, eliece77}. 
Note that with our approach, we avoid the greedy research of the set $\mathbf{U}$ among all the possible $\binom{m}{k}$ solutions which realizes $\mathbf{H}( \mathbf{y}| \mathbf{U}) \rightarrow 0$ .
Instead, we adopt the stochastic approach. Precisely, we associate each $i$-th feature with a random variable $z_i\sim \mathrm{Bernoulli}(p_i)$. The CE-based algorithm identifies which variables  $z_i, \: i=1,\cdots,m$ must have $p_i \rightarrow 1$, so that the objective function $\mathcal{O}(\mathbf{U}(\mathbf{z})) = \mathbf{H}( \mathbf{y}| \mathbf{U})$ gets close to $0$. This is called \ac{ASP}  \cite{rubinstein04}. In this way, we get the optimal distribution of the binary vector $\mathbf{z}$ through which we identify the features to be selected, i.e. the $i$-th feature is selected if $p_i \rightarrow 1$. It is worth noting that searching for the solution of the optimization problem through the definition of the \ac{ASP} has the advantage of addressing the native  problem in (\ref{Native Optimization Porblem}) as a convex problem.\footnote{More details are in Section 4 of \cite{rubinstein04}.} 

We formulate the \ac{ASP} as a minimization problem, as shown in Eq. \ref{parameter estimation}. In the following we present the essential steps that brings to its formulation.  
Briefly, we need to find the probability distribution $g(\mathbf{z},\mathbf{p})$ of the values in $\mathbf{z}$ equal to $1$ that solves the equation: 
%
%
\begin{equation} 
  \label{probability estimation}
         \mathrm{Pr}(\mathcal{O}(\mathbf{U}(\mathbf{z}))\leq \gamma)=\sum_{\{\mathbf{z}\}}\mathcal{I}{(\mathcal{O}(\mathbf{U}(\mathbf{z}))\leq \gamma)} \:g(\mathbf{z},\mathbf{p})
 \end{equation}%
where  $\mathcal{I}{(\cdot)} $ is the indicator function of the event $\mathcal{O}(\mathbf{U}(\mathbf{z}))\leq \gamma$, and $ \gamma$ is the minimum value for our objective function. Precisely, $\gamma$ at step t is calculated as the percentile $1-\beta$ of the objective function calculated by using the samples drawn from the distribution $g(\mathbf{z},\mathbf{p})$ at step t. Note that, the authors in \cite{rubinstein04} recommend to set $\beta$ in the range $0.9-0.95$. The indicator function is equal to $1$ for all the possible configurations in $\mathbf{z}$ that  verify the event $\mathcal{O}(\mathbf{U}(\mathbf{z}))\leq \gamma$, and $0$ otherwise.

We estimate $g(\mathbf{z},\mathbf{p})$ through the Likelihood Ratio (LR) estimator with reference parameter $\mathbf{p}$. Precisely, we apply the LR theory of estimation \cite{rubinstein04} to define the following optimization problem and to obtain the optimal value $\mathbf{p}^*$ for the distribution. 
\begin{equation} 
  \label{parameter estimation}
         \mathbf{p}^*=\displaystyle arg\min_{\mathbf{p}} \frac{1}{S} \sum_{j=1}^{S}\mathcal{I}{(\mathcal{O}(\mathbf{U}(\mathbf{z}_j))\leq \gamma)} \:\mbox{ln}(g(\mathbf{z}_j,\mathbf{p}))
 \end{equation}%
where $\mathbf{Z}=\{\mathbf{z}_1, \cdots, \mathbf{z}_S\}$ is a set of possible  samples drawn from the distribution $g(\mathbf{z},\mathbf{p})$.

As stated above $\mathbf{z}_j=[z_{1j} \cdots z_{mj} ]$ is a vector of independent Bernoulli random variables where $z_{ij}$ takes value equal to $1$ with probability $p_{i}$ and $0$ with probability $1-p_{i}$. 
Hence, $g(\mathbf{z}_j,\mathbf{p})$ 
can be written as:

\begin{equation} 
  \label{pdf}
                           g(\mathbf{z}_j,\mathbf{p})=\displaystyle \prod_{i=1}^m p_i^{z_{ij}}(1-p_i)^{(1-z_{ij})} \: ; \: z_{ij} \in \{0,1\}
 \end{equation}
 
Given that the objective function of problem (\ref{parameter estimation}) is concave\footnote{The logarithm is a concave function, the indicator function is $0$ or $1$ so the weighted sum of concave functions gives still a concave function.}, we can solve it in closed form by imposing: $$\frac{\partial{}}{\partial{p_i}}\frac{1}{S} \sum_{j=1}^{S}\mathcal{I}{(\mathcal{O}(\mathbf{U}(\mathbf{z}_j))\leq \gamma)} \:\mbox{ln}(g(\mathbf{z}_j,\mathbf{p}))=0,$$ leading to: 
\begin{equation} 
p_i=\frac{\sum_{j=1}^{S}\mathcal{I}{(\mathcal{O}(\mathbf{U}(\mathbf{z}_j))\leq \gamma)}z_{ij}}{\sum_{j=1}^{S}\mathcal{I}{(\mathcal{O}(\mathbf{U}(\mathbf{z}_j))\leq \gamma)}}\:\;\: i=1 \cdots m;
\label{closed form solution}
\end{equation}

In the CE-base algorithm the result in the equation (\ref{closed form solution}) is used for updating the distribution $\mathbf{p}$ as follows:

\begin{equation}
\label{closed form solution2}
p_i=(1-\alpha)p_i+\alpha \frac{\sum_{j=1}^{S}\mathcal{I}{(\mathcal{O}(\mathbf{U}(\mathbf{z}_j))\leq \gamma)}z_{ij}}{\sum_{j=1}^{S}\mathcal{I}{(\mathcal{O}(\mathbf{U}(\mathbf{z}_j))\leq \gamma)}} .
\end{equation}
The mathematical analysis about the choice of the parameter $\alpha$ is provided in the Appendix\ref{appendixa} of this work. Further indications on the choice of $\alpha$ can be found in \cite{rubinstein04,Costa2007, Wu2014}.
The derivation of equations (\ref{parameter estimation}-\ref{closed form solution}), as well as, the optimality of $g(\mathbf{z}_j,\mathbf{p})$ are proven in \cite{rubinstein04}.


The solution of the problem defined in (\ref{parameter estimation}) is achieved through Algorithm \ref{algo:main_CE}: it starts with an initial guess of $\mathbf{p}_G$; $S$ Bernoulli random samples of size $m$ each (line 4) are drawn at each step $t$. For each sample $\mathbf{z}_s$, the values of the conditional entropy (line 7) are computed on the dataset where the only active features are those corresponding to the elements equal to one (line 6) in $\mathbf{z}_{s}$. The subset selection is shown in the procedure $\mathrm{\textsc{getSubset}}(\mathbf{X},\mathbf{z})$ (lines 15-26). Then we compute $\mathbf{p}(\mathbf{Z}_t)$ (lines 9-10) as in Eq. (\ref{closed form solution}) and finally we update the current estimate of the probability vector $\mathbf{p}$ (line 11) as in Eq. (\ref{closed form solution2}). 

\begin{algorithm} [ht]
\caption{CE-based algorithm for FS}
\label{algo:main_CE}
\begin{algorithmic}[1]
\Procedure{CE}{$\mathbf{X},\mathbf{y},\mathbf{p}$, T,S}
\ForAll {$t=1,\dots, T$}
\State $\mathbf{Z}_t\gets$\textsc{genRndSample}$(S,\mathbf{p})$
\Comment{$\mathbf{Z}\in\{0,1\}^{S\times m}$}
\State $\mathbf{u}\gets \{\}$
\ForAll{$\mathbf{z}_s \in \mathbf{Z}_t$}
\Comment{$\mathbf{z}_s \in \{0,1\}^{1\times m}$}
\State $\mathbf{U} \gets$ \textsc{getSubset}($\mathbf{X}$,$\mathbf{z}_s$)
\State $\mathbf{u}\gets\mathbf{u}\cup \mathbf{H}(\mathbf{y}|\mathbf{U})$
\EndFor
\State $\gamma \gets$ \textsc{computePercentile}($\mathbf{u},1-\beta$)
\State $\mathbf{p}(\mathbf{Z}_t) \gets $\textsc{computeNewProb}($\mathbf{u},\gamma,\alpha$) 
\Comment{Eq. (\ref{closed form solution})}
\State $\mathbf{p} \gets (1-\alpha) \mathbf{p} + \alpha \mathbf{p}(\mathbf{Z}_t)$
\Comment{Eq. (\ref{closed form solution2})}

\EndFor

\State \Return $\mathbf{p}$
\EndProcedure
\Procedure{getSubset}{$\mathbf{X}$,$\mathbf{z}$}
\State $\mathbf{U} \gets \{\}$
\ForAll{$\mathbf{x} \in \mathbf{X}$}
\State{$\mathbf{u} \gets \{\}$}
\ForAll{$j=1,\dots,m$}
\If{$z_j == 1$}
\State $\mathbf{u} \gets \mathbf{u} \cup x_{j}$
\EndIf
\EndFor
\State $\mathbf{U} \gets \mathbf{U} \cup \mathbf{u}$
\EndFor
\EndProcedure
\end{algorithmic}
\end{algorithm}

\section{Federated Feature Selection}
\label{sec:fed_approach}
In this section we present how we exploit the CE-based \ac{FS} algorithm presented in Section \ref{sec:FE_theory_MAIN} and summarised in Algorithm \ref{algo:main_CE} to design our \ac{FFS} algorithm \ac{FFS}, described in Algorithms \ref{algo:federated_algo_server} and \ref{algo:federated_algo_client}. They cover, respectively, the two functional blocks of \ac{FFS}, i.e., Algorithm \ref{algo:federated_algo_server} is executed by the \ac{ES}  to coordinate the distributed \ac{FS} and Algorithm \ref{algo:federated_algo_client} runs on the clients. The \ac{FFS} is an iterative procedure. At the beginning, the \ac{ES} sends to the clients involved in the process a vector $\mathbf{p}_G \in \mathbb{R}^m$ where each element represents the probability that each feature has to be selected according to its importance (lines 8-10 of Alg.\ref{algo:federated_algo_server}). Each element of $\mathbf{p}_G$ is initialized to $0.5$, i.e., this is a common choice when using the CE algorithm. The vector $\mathbf{p}_G$ represents a piece of global information that the \ac{ES} shares with the client nodes.  Each client $l$ uses $\mathbf{p}_G$ to initialize its local copy of the probability vector, i.e., $\mathbf{p}_l\gets\mathbf{p}_G$ and runs the local \ac{FS} procedure based on its local data (lines 2-3 of Alg. \ref{algo:federated_algo_client}). At the end of the local \ac{FS}, the $l$-th client sends to the \ac{ES} the locally updated probability vector  $\mathbf{p}_{l_{new}}$ and a control information regarding the cardinality of its local data $n_{l}$ whose purpose will become clear in the following. The \ac{ES} computes the new global probability vector (line 13 of Alg. \ref{algo:federated_algo_server}) by aggregating the ones received from the clients as expressed in Equation (\ref{eq:joint_probability}) and discussed later on. The updated vector $\mathbf{p}_G$ is transmitted to the nodes that run Algorithm \ref{algo:federated_algo_client} by updating the local probability vector with the new global one. This procedure iterates until the distribution global probability vector converges to a stable one. In \ac{FFS} we check convergence by comparing the distribution of the current global probability vector $\mathbf{p}_G$ to the previous one $\mathbf{p}_{G_{old}}$ using the Kolmogov-Smirnov statistical test for two one-dimensional samples (KS-test). The procedure stops when \textit{(i)} the p-value of the KS-test is greater than a fixed threshold\footnote{We empirically observed that the closer $\tau_1$ to one, the more accurate the solution.} $\tau_1=0.995$ and, \textit{(ii)} its variation from the previous one is less than $\tau_2 = 10^{-6}$ (line 7 of Alg. \ref{algo:federated_algo_server}).

The core point of Algorithm \ref{algo:federated_algo_server} regards the aggregation step (line 13 of Alg. \ref{algo:federated_algo_server}) where the \ac{ES} merges the local probability vectors into the global one which, in our solution, is defined as a weighted average. The main idea is to merge the local probability vectors by a weighted average where the weights (computed as in Eq. (\ref{eq:probability_data_from_node})) serve the twofold purpose of \textit{(i)} considering more (or less) those vectors that are computed from larger local datasets and \textit{(ii)} defining a common support among all the probability vectors. This second aspect is quite crucial for the consistency of the computation in Eq. (\ref{eq:joint_probability}).

Formally, we assume that each node acquires a number of \textit{i.i.d.} records $n_{l}$  to perform the \ac{FS}, and that the nodes share the same set of features $\mathbf{X}$.
The global probability $\mathbf{p}_G$ used for the \ac{FS}  can be written as follows:
\begin{equation}
\label{eq:joint_probability}
    \mathbf{p}_{G}=\sum_l \mathbf{p}_{l} \omega_{l},
\end{equation}
 where $\mathbf{p}_{l}$ is the solution of problem (\ref{parameter estimation}) at node $l$ obtained by using Algorithm \ref{algo:main_CE},
 and $\omega_l$ weights $\mathbf{p}_l$ \textit{w.r.t.} the other nodes, whose formal definition is:
\begin{equation}
    \label{eq:probability_data_from_node}
    \omega_{l}=\frac{n_{l}}{\displaystyle\sum_l n_{l}}.
\end{equation}

As anticipated, according to equation (\ref{eq:probability_data_from_node}), we weight the probability vector $\mathbf{p}_l$ of node $l$ proportionally to the size of its local dataset compared to the whole amount of data present in the system. In this way, we can contrast situations where local datasets are heterogeneous w.r.t. the size.

In \ac{FFS}, the updating scheme can be, at least in principle, both synchronous and asynchronous, provided that the set of nodes involved in the process does not change over time.\footnote{Note that this condition does not imply that all nodes must be active during the entire process. In fact, as we will show in Section \ref{sec:num_results} our system is robust to the presence of churning nodes.} Precisely,  we  assume  a  system  where the  \ac{ES} after having sent the updated global probability vector, expects the nodes to receive their local updates within a fixed time slot, after which, it begins the aggregation step  using only the information received. Therefore, the number of updates used to compute the new global probability vector might change because a subset of nodes could not communicate their updates within the deadline set by the \ac{ES}. Regardless of the number of nodes that contributed to the aggregation step during one round of communication, the \ac{ES} broadcasts the new global probability vector $\mathbf{p}_G$ to \emph{all} nodes in the system. In this way, all nodes start the new round of local computation from the same starting point, and, consequently, we dramatically limit the potentially detrimental effects deriving from the aggregation of outdated local probability vectors. Moreover, as proved by the convergence analysis provided in Appendix\ref{appendixa} and Appendix\ref{appendixb}, independently from the updating scheme, \ac{FFS} converges in a finite number of steps to the very same solution as running the CE in centralised settings i.e., with complete access to the entire dataset.

It's worth noting that our solution is able to cope with feature redundancy in federated settings. Precisely, this represents an issue that might prevent the possibility of performing the \ac{FS} in federated settings. In fact, running a standalone \ac{FS} algorithm on different local datasets where there is redundancy between features, different \acp{FS} might occur but with an equivalent information content across all the \ac{AV}s. This aspect makes all the local selections completely useless regarding the communication efficiency, due to the consequent lack of agreement on the \ac{FS} between the \ac{AV}s. Conversely, since in \ac{FFS} the \ac{AV}s share at each communication round their local information, they may come up with a final agreement on the \ac{FS}. Summarising, even if there is redundancy between features, the final selection is consistent among all the AVs and, according to results presented in Section \ref{sec:num_results}, it is also accurate if compared to the centralized \ac{FS} (i.e., when all the local raw data are transferred onto the \ac{ES}).

\begin{algorithm} [ht]
\caption{Server side \ac{FFS} algorithm}
\label{algo:federated_algo_server}
\begin{algorithmic}[1]
\Procedure{Server-Node}{}
\State $v \gets 0$
\Comment{p-value of Kolmogorov-Smirnov test}
\State $\tau_1 \gets .995$
\State $\tau_2 \gets 10^{-6}$
\Comment{Thresholds to check convergence}
\State $\mathbf{p}_{G} \gets \{1/2\;|\;\forall\; p_{i}\;  i=1,\dots,m\}$
\Do
\ForAll {$l\in L$}
\State \textsc{sendToClient}($l,\mathbf{p}_{G}$)
\EndFor
\State \textsc{receiveFromClients}($\mathbf{p}_{l_{new}},n_{l}$)
\State $\mathbf{p}_{G_{old}} \gets \mathbf{p}_{G}$
\State $\mathbf{p}_{G} \gets$ \textsc{updateGlobalProbability}()
\Comment{Eq. (\ref{eq:joint_probability})}
\State $v_{old} \gets v$
\State $v \gets $\textsc{KolmogorovSmirnovTest}$(\mathbf{p}_{G},\mathbf{p}_{G_{old}})$
\doWhile{$v \geq \tau_1 \land |v-v_{old}| \leq \tau_2$}
\Comment{repeat until convergence is met}
\EndProcedure
\end{algorithmic}
\end{algorithm}

\begin{algorithm} [ht]
\caption{Client side Federated Feature Selection algorithm}
\label{algo:federated_algo_client}
\begin{algorithmic}[1]
\Procedure{Client-Node}{}
\State $\mathbf{p}_l \gets$ \textsc{receiveFromServer}($\mathbf{p}_G$)
\State $\mathbf{p}_{l_{new}} \gets$ CE($\mathbf{X}_l,\mathbf{y}_l,\mathbf{p}_l$,T,S)
\Comment{Algorithm \ref{algo:main_CE}}
\State \textsc{sendToServer}($\mathbf{p}_{l_{new}}, n_l$)
\EndProcedure
\end{algorithmic}
\end{algorithm}

\section{Numerical evaluation}
\label{sec:num_results}
In this section, we present the numerical results of our compression method based on the \ac{FFS} algorithm presented in Section \ref{sec:fed_approach}.
Before going through the results, we introduce the datasets, the simulation settings, the methodology, and the metrics used to evaluate our solution's performance.
 
\subsection{Dataset description and simulation settings}
We based the performance evaluation of \ac{FFS} on two datasets, each one mapping one of the two use cases described in Section~\ref{sec:problem_statement}. The first one called MAV\footnote{dataset available at: https://tinyurl.com/mavmr01} is a publicly available dataset containing both 64$\times$64 images and $6$ \acp{IMU} collected by a \ac{AV} during a mission in a controlled environment. The second dataset called WEarable Stress and Affect Detection (WESAD) is a collection of data sampled from heterogeneous biophysical sensors: ECG, EDA, EMG, Temperature, Respiration and Inertial Measurements on the three axes. 

\paragraph{MAV dataset}
both images and inertial measurements are synchronised to obtain a set of eRIDs. We pre-process the raw images to extract more informative features as it is customary in the computer vision domain. Feature extraction eases the training of a machine learning model and, performs a preliminary step of data compression. In fact, a raw image is made of 4102 floats (64$\times$64 pixels + 6 IMU readings) while, after the feature extraction, it shrinks down to a vector of size 2166 floats. In our settings, we extract the \ac{HOG} features\footnote{HOG is a standard feature extraction methodology used in computer vision and image processing to create an image descriptor that captures the spatial relations between different portions of it \cite{chen14}. }, and we assume that the feature extraction is accomplished directly on the \ac{AV}, which might be possible if equipped with a board of the kind discussed in \cite{chen14}.
Note that the original dataset is unlabeled. Therefore we labelled it in a way compatible with the original context of positioning. To this end, we associated with each eRID a label corresponding to the corresponding voxel.\footnote{A voxel represents a value on a regular grid in three-dimensional space.}
Table \ref{tab:erid-mav} shows the structure of an eRID for the MAV; the first 2160 feature are HOG while the last 6 are IMUs, i.e., acceleration (ACC) and angular velocity (AV). 
\begin{table}[ht!]
    \centering
    \caption{Structure of a MAV eRID}
    \label{tab:erid-mav}
    \begin{tabular}{|c|c|c|c|c|c|}
    \hline
    0 & 1 & 2 & ... &  2158 & ... \\
    \hline
    \multicolumn{6}{|c|}{HOG \#} \\
    \hline \hline
      2160 & 2161 & 2162 & 2163 & 2164 & 2165  \\
    \hline
     ACC$_x$ & ACC$_y$ & ACC$_z$ & AV$_x$ & AV$_y$ & AV$_z$ \\
    \hline
    \end{tabular}
    
\end{table}
The whole dataset contains 2911 labelled records. To simulate the federated data collection, we split it into $10$ disjoint partitions of size 291 records such that each partition is i.i.d. w.r.t. the entire dataset. Each subset represents a \ac{AV}. The data collection is slotted; hence, the \acp{AV} draw with replacement a random sample from their local dataset for each time slot. This sample is used to perform the local computation of the distributed algorithm followed by a communication round for synchronising the \acp{AV} on the local \ac{FS}. Each random draw's size is accumulated to trace the cache necessary for storing data until the completion of the distributed \ac{FS}.

\paragraph{WESAD dataset} it provides data in terms of features and labels already useful to perform the detection of stress and affection state of human subjects. The dataset contains readings from two devices, i.e., Respiban and Empatica E4, positioned i) on the chest and ii) on the wrist of human subjects. Each device is equipped with multiple sensors monitoring several physiological parameters. Since the two devices have different operating settings, we focused on the Respiban, whose collection rate is homogeneous for all its sensors. 
The dataset contains readings collected from 17 human subjects, which perform a predetermined protocol to induce the body in one of the following states: 0-baseline, 1-amusement, 2-stress, 3-meditation, 4-recovery. The data collected for each subject amounts to $\sim$3.6M records, equivalent to $\sim$220 MB. A complete description of the dataset is provided in \cite{Schmidt2018}.
Table \ref{tab:erid-wesad} shows the structure of an eRID for the WESAD.
\begin{table}[ht!]
    \centering
    \caption{Structure of a WESAD eRID}
    \label{tab:erid-wesad}
    \begin{tabular}{|c|c|c|c|c|c|c|c|}
    \hline
    0 & 1 & 2 & 3 & 4 & 5 & 6 & 7  \\
    \hline
    ACC$_x$ & ACC$_y$ & ACC$_z$ & ECG & EMG & EDA & TEMP. & RSP \\
    \hline
    \end{tabular}
\end{table}
Due to the huge size of the dataset we used the data from 5 out of 17 subjects, corresponding to $\sim$1.1 GB. The data is already partitioned   according the subject ID, thus we keep the original partitions. In our simulated scenario, each partition corresponds to an edge device holding the data of only one subject, i.e., no artificial data re-distribution is performed. As for the previous scenario, each device executes \ac{FFS} using only its own data.

We evaluate the performance of our methodology according to two metrics: 
\begin{itemize}
    \item \emph{accuracy}: to assess the quality of the distributed \ac{FS}
    \item \emph{network overhead ($N_{OH}$)}: to evaluate the impact in terms of network traffic generated by our methodology
\end{itemize} 
Our target is to compress the data to be transmitted, without significantly degrading its informative content.

\paragraph*{Accuracy metric} The quality assessment is a two-stage procedure. First, we set the baseline validating the quality of the features selected by CE executed in a centralised setting, i.e., we train a classifier using the set of selected features (CE-CFS) on the entire dataset, and we compare its prediction performance with that of a second classifier trained on the whole set of features (NO-FS). If the CE-CFS performance on a smaller group of features is comparable or equivalent with the one identified by NO-FS, we consider the \ac{FS} valid. To strengthen this initial evaluation, we compare the centralised results of CE-CFS with other three reference \ac{FS} algorithms: mRMR\cite{mrmr}, HJMI\cite{hjmi} and ANOVA\cite{heiberger2009one}. As we will show in the following, for all these benchmarks we have to specify the size $k$ of the features selection. Since we are interested in assessing the quality of the \ac{FS} and for the sake of fairness, we set $k$ equal to the size of the \ac{FS} obtained by CE-CFS (which finds such a number in a completely autonomous way).
    
Then, we repeat the same procedure training another classifier on the subset of features obtained from our \ac{FFS} and we compare its performance with all the centralised methods. 
We split the dataset in train (80\%) and test set (20\%). The train set is used for both \ac{FS} and model training, while the test is used for performance evaluation only. 
The accuracy is defined as the average of correctly classified records 
\begin{equation}
    A = \frac{1}{N}\sum_{i=1}^N I(\hat{y}_i = y_i),
\end{equation}
where N is the size of the test set, $I$ is the indicator function, $\hat{y}_i$ and $y_i$ are the $i$-th predicted and true label, respectively. 
For the sake of statistical significance, the training is repeated ten times, changing the initialisation of the classifier and the composition of training and test set. The reported results are average values accompanied by confidence intervals at 95\%. 

\paragraph*{Network Overhead} we measure the network traffic generated by our solution as follows. 
On the one hand, we compute the network overhead generated by the \ac{FFS} network defined as: 
\begin{equation}
    N_{OH} = R * L * 2*(z+1+b)
\end{equation}
where $R$ is the number of communication rounds before all the $L$ \acp{AV} involved in the distributed \ac{FS} converge to a solution, $z+1$ is the number of nonzero floating point numbers belonging to the probability vector $\mathbf{p}_l$ in \eqref{eq:joint_probability} exchanged between the \acp{AV} during each round plus the weight $\omega_l$ 
in \eqref{eq:probability_data_from_node}. The symbol $b$ is the size of the bit map used to reconstruct the position of the non-zero elements exchanged between the \acp{AV} and the edge server. 
On the other hand, we compute the compression obtained through the \ac{FS} as: 
\begin{equation}
    C = |F| / |D|
\end{equation}
where $F\subseteq D$ is the selected set, and $D$ is the entire set of features. 

\subsection{Settings the baseline: FS in centralised settings}
\label{ssec:accuracy}

\begin{figure*}[ht]
    \centering
    \subfloat[C-HOG]{
    \centering
    \includegraphics[width=.35\linewidth]{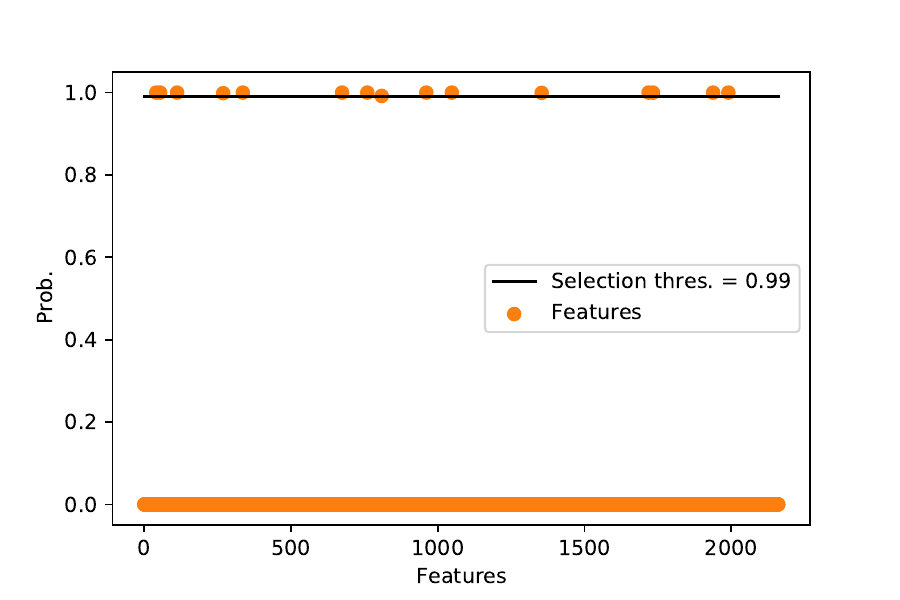}
    \label{fig:gprob_hog}
    }
    \subfloat[C-IMU]{
    \centering
    \includegraphics[width=.35\linewidth]{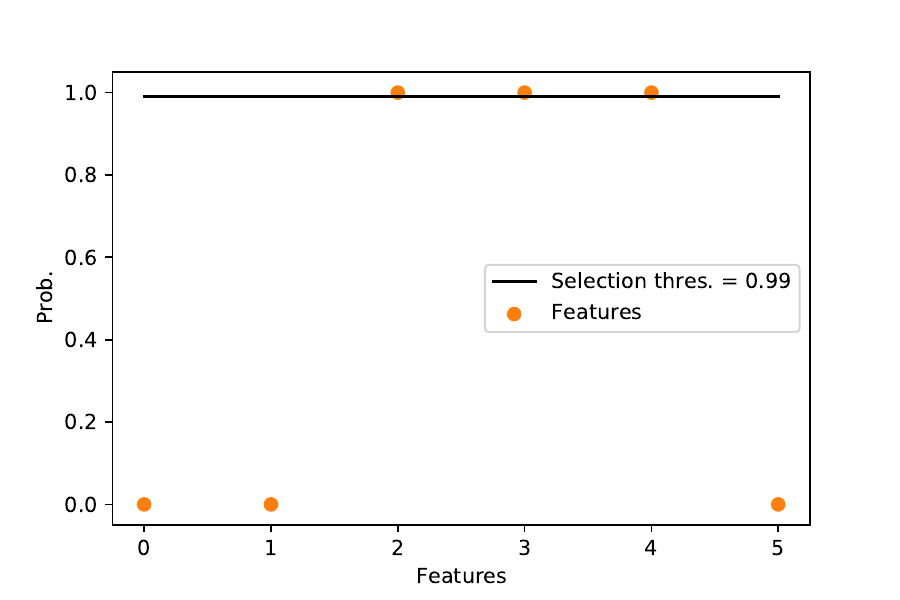}
    \label{fig:gprob_imu}
    }
  \caption{Centralised \ac{FS} probability for HOG and IMU. The selected features are those with probability greater than 0.99 (above threshold). }
    \label{fig:gprob_fs}
\end{figure*}

The following results regard the first stage of the validation, i.e., the accuracy of a classier trained using only the subset of features identified by the CE algorithm w.r.t  the performance obtained by a classifier trained  on the entire dataset. For this stage of validation, we train a Neural Network (NN). For MAV the NN is a multi-layer perceptron with two hidden layers of 300 and 100 neurons each.  For WESAD, we used a deep NN with four hidden layers of 300,100,64,32 neurons each. The input layer's size depends on the number of features selected, while the size output layer is 37 and 5 for MAV and WESAD, respectively. The activation function is ``ReLU''\footnote{REctified Linear Unit} and the optimizer is ``Adam''\footnote{Stochastic Gradient Descent with ADAptive Momentum} for both the models. These are very common settings which typically provides good performance~\cite{goodfellow2016}.

        
\begin{table}[ht]
    \centering
    \caption{Comparison between NO-FS and CE-CFS on MAV and WESAD dataset.}
    \begin{tabular}{llllll}
    \toprule
        Dataset & Method  & Size & FS & $C$ & Accuracy  \\
          & & (\# record) & (\#) & (\%) & (\%) \\
          \midrule
          \multirow{5}{*}{MAV} & NO-FS & 2911 & 2166 (All) & - & 97.5$\pm$0.4 \\
            & CE   & 2911 & 18 &  99 &96.7$\pm$0.5  \\
            & MRMR   & 2911 & k=18 &  99 &95.0$\pm$0.5  \\
            & ANOVA   & 2911 & k=18 &  99 &95.0$\pm$0.4  \\
            & HJMI  & 2911 & k=18 &  99 &96.3$\pm$0.7  \\
            \\
          \multirow{5}{*}{WESAD} & NO-FS & 15$*10^6$ & 8 (All) & - & 94.3$\pm$0.7 \\
            & CE & 15$*10^6$ & 4 & 50 & 94.6$\pm$0.8 \\
            & MRMR    & 15$*10^6$ & k=4 &  50 &94.3$\pm$0.8  \\
            & ANOVA   & 15$*10^6$ & k=4 &  50 &94.5$\pm$0.5  \\
            & HJMI   & 15$*10^6$ & k=4 &  50 &90.2$\pm$1.6  \\
      \bottomrule
    \end{tabular}
    \label{tab:cvalid}
\end{table}
Results in Table \ref{tab:cvalid} show that CE algorithm executed on both datasets in centralised settings can autonomously identify a minimal set of features (i.e., 18 for MAV and 4 for WESAD) with the very same informative content of the whole feature set.  The accuracy obtained by both the NN models trained on the CE's \ac{FS} is statistically equivalent to the one obtained on the whole set of features, inducing a quite impressive compression rate ($C$): up to 99\% and 50\% of network traffic for MAV and WESAD, respectively. As a further confirmation of the CE results, we perform the \ac{FS} using other three reference benchmarks, i.e., MRMR, ANOVA, HJMI. Note that all these approaches select a subset of features with the very same informative content of CE. However, we point out that for all of them we have to decide beforehand the number of features to be selected. This represent a major shortcoming that, instead, CE-based methods overcome by design, since the number of features to be selected is a byproduct of the CE algorithm.  Finally, these results assess the suitability of the CE algorithm on both datasets, thus we can use them as a benchmark for the evaluation of our distributed \ac{FFS} method. 

\begin{figure*}[ht]
    \centering
    \subfloat[F-HOG]{
    \centering
    \includegraphics[width=.35\linewidth]{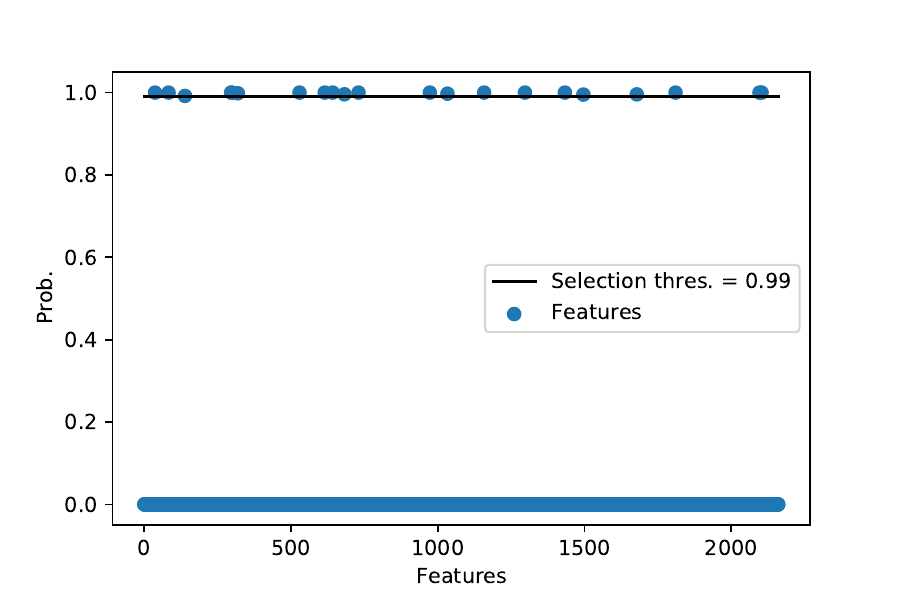}
    \label{fig:fprob_hog}
    }
    \subfloat[F-IMU]{
    \centering
    \includegraphics[width=.35\linewidth]{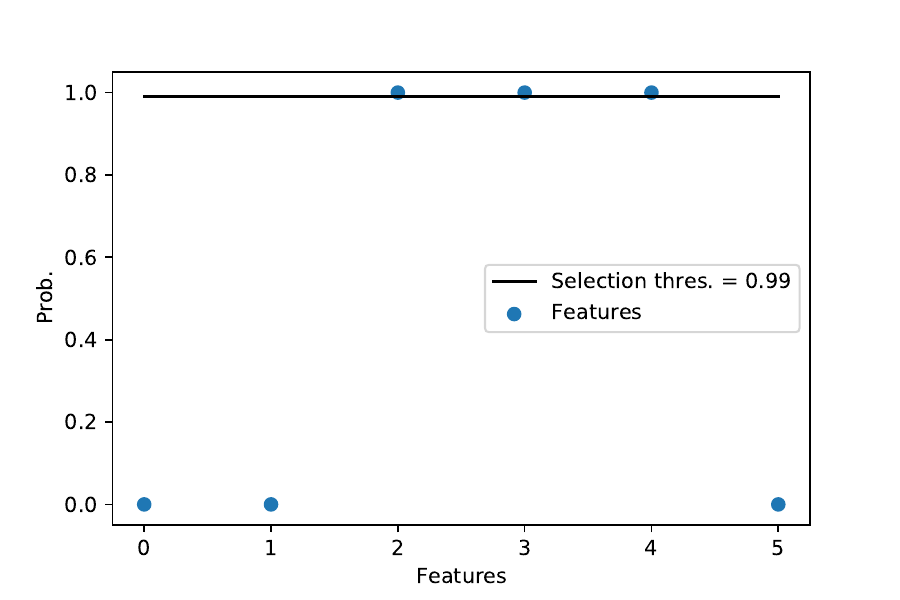}
    \label{fig:fprob_imu}
    }
    \caption{\ac{FFS} probability for HOG and IMU. The selected features are those with probability greater than 0.99 (above threshold). }
    \label{fig:fprob_fs}
\end{figure*}

\subsection{Evaluation of Federated Feature Selection}
We focus now on the analysis of our FFS method. We compare its performance to those obtained by CE executed in centralised settings (CE-CFS). We recall that, in federated (distributed) settings, each \ac{AV} can process only the data it locally collects.

First we assess the performance of FFS in a static distributed scenario where the \acp{AV} have collected all the data and, before sending them to the \ac{ES}, they perform the distributed \ac{FS} in order to transmit only the very necessary information. 
\begin{table}[ht]
    \centering
 \caption{Comparison between CE-CFS and FFS on MAV and WESAD.}
    \begin{tabular}{llllll}
        \toprule
        Dataset& Method & Size & FS & $C$ &  Accuracy \\
        &  & (\# obs.) &(\#) & (\%)&  (\%$\pm$C.I.) \\
        \midrule
         \multirow{2}{*}{MAV}& CE-CFS & 2911 & 18 & 99  & 96.7$\pm$0.5 \\
         & FFS & 291 & 24 & 99  &96.7$\pm$0.4 \\
         \\
         \multirow{2}{*}{WESAD}& CE-CFS & 15M & 4 & 50  & 94.6$\pm$0.8 \\
         & FFS & 3M & 4 & 50  &94.6$\pm$0.8 \\
         
         \bottomrule
    \end{tabular}
    \label{tab:mav-cfsvsffs}
\end{table}

Table~\ref{tab:mav-cfsvsffs} reveals that for MAV dataset, \ac{FFS} finds a set of features that, although slightly larger than that found by CE-CFS (24 instead of 18), it has the very same informative content, i.e., the accuracy of the NN model trained on both subsets of features are statistically equivalent. 
As we can see, the results also hold for the WESAD dataset. Precisely, \ac{FFS} selects the same number of features identified by CE-CFS. Specifically, \ac{FFS} and CE-CFS select the same set, i.e., the features with indexes [1,2,5,6], explaining why the NN achieves the same prediction accuracy. We motivate such an exact correspondence between \ac{FFS} and CE-CFS selection considering that the small size of the complete feature set of WESAD might prevent a high number of feature subset with equivalent informative content. Such an assumption also holds for the MAV dataset. In fact, as we can see in Figures~\ref{fig:fprob_hog}, \ref{fig:gprob_imu} \ac{FFS} and CE-CFS select the same subset of IMU features. Conversely, when the set of features is more redundant, as it is for the HOGs, there might exist several subsets holding the same informative content. The comparison in Figures ~\ref{fig:fprob_hog} and~\ref{fig:gprob_hog} confirms such a claim because the two feature sets are overlapping but not equal, yet the overall accuracy is comparable.

\begin{figure}[ht]
    \centering
    \includegraphics[width=.8\linewidth]{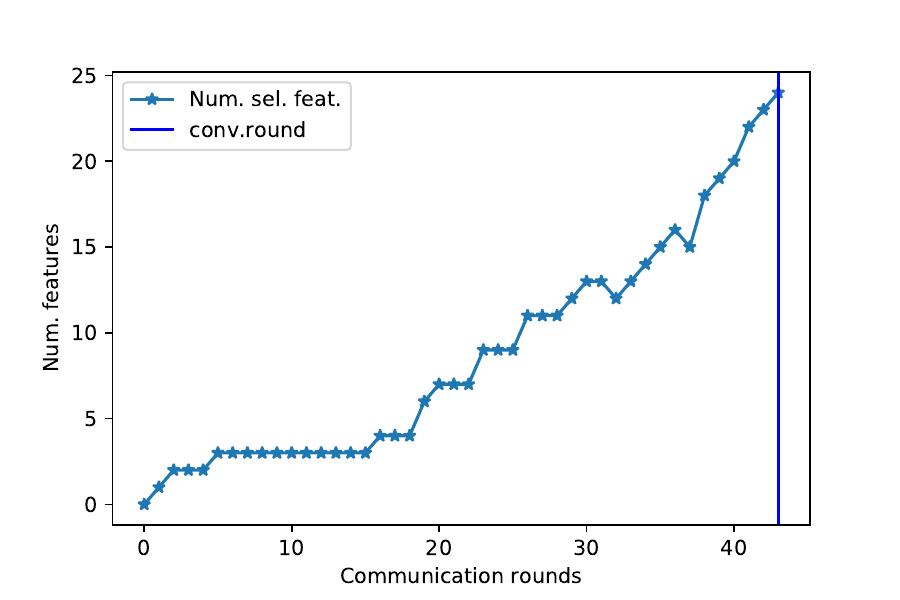}
    \caption{\ac{FFS} process on MAV dataset.}
    \label{fig:f-sel}
\end{figure}

This result provides a preliminary insight regarding the effectiveness of the Bayesian aggregation used to merge the information extracted by the \acp{AV} from their local datasets. Precisely, Figure~\ref{fig:f-sel} shows the number of selected features at each communication round for the MAV case. As we can see, in the beginning, the cardinality of \ac{FS} remains almost constant. In this phase, due to the partitioning of data in separated datasets, the CE algorithm has not yet enough knowledge to identify the most informative features. However, the number of features added to the selection starts increasing following an almost-linear trend in a few communication rounds (16). The process ends after 44 rounds, i.e. when the distribution of probabilities indicating the most informative features becomes stable.

\begin{table}[ht]
    \caption{Performance of \ac{FFS} varying the data processed during a communication round}
    \begin{tabular}{lllllllll}
        \toprule
        Dataset  & Size & FS & Accuracy  & $R_{c}$ & $C$ & $N_{OH}$ & Cache \\
        &(\# obs) & (\#) & (\%$\pm$C.I.) & (\#) & (\%) & (MB) & (MB)  \\
        \midrule
        \multirow{5}{*}{MAV} & 291 & 24 &96.7$\pm$0.4 & 44 & 99 & 16 & 217 \\
        & 203 & 26 & 96.5$\pm$0.3 & 37 & 99 & 13 & 128 \\
        & 145 & 34 & 97.0$\pm$0.3 & 55 & 98 & 20 & 134 \\
        & 87 & 59 & 97.2$\pm$0.3 & 43 & 97 & 15 & 63 \\
        & 29 & 41 & 97.2$\pm$0.4 & 53 & 98 & 19 & 26 \\
        \\
         \multirow{2}{*}{WESAD} & 3$\cdot10^6$ & 4 & 93.6$\pm$0.8 & 11 & 50 & 0.009 & 2$\cdot10^3$   \\
         & 1$\cdot10^6$  & 5 & 93.8$\pm$0.5 & 10 & 38 & 0.008 & 1$\cdot10^3$  \\
        \bottomrule
    \end{tabular}
    \label{tab:net_perf}
\end{table}

Our method's capability to converge quickly to the final and most informative set of features directly affects the amount of network traffic generated upon the completion of the \ac{FFS}. To confirm such a claim, we performed a set of simulation in which we run  \ac{FFS} varying the size of the local dataset available at the edge device. In this way, we want to analyse our method's robustness when each edge device can access only a limited amount of data. In Table~\ref{tab:net_perf} we report the size of data used for each update (Size), the number of selected features (FS), the accuracy, the number of communication rounds upon convergence ($R_c$), the compression obtainable with \ac{FFS} ($C$), the network overhead generated by \ac{FFS} ($N_{OH}$), and the size of the cache needed to collect the data before starting the data transmission. 
Overall, we observe that, for both datasets, decreasing the size of data processed at each round does not affect significantly the number of communication rounds needed by \ac{FFS} to converge to a solution, which results in limiting the network overhead generated during the process. Specifically, considering a dynamic data collection process as in the MAV-related use case, we see that the network overhead is always  i) less than the storage needed to cache the data before starting the transmission and ii) negligible considering the compression achieved (i.e., up to 99\%). 
Interestingly, the same holds also for the WESAD scenario. In this case, the network overhead can be considered negligible w.r.t. the size of the data processed ($<1$MB) if compared with the compression rate achieved by \ac{FFS} (up to 50\%). 

\begin{table}[ht!]
    \centering
    \caption{Performance of \ac{FFS} varying the percentage non-faulty \acp{AV} per communication round}
    \begin{tabular}{lllllll}
    \toprule
    Dataset &  $\rho$  & FS & Accuracy & $R_c$ & $C$   \\
     & & (\#) & (\%$\pm$C.I.) & (\#) & (\%)    \\
    \midrule
    \multirow{2}{*}{MAV} & 0.2 & 48 & 97.0$\pm$0.4 & 37 & 98  \\
    & 0.3  & 80 & 97.2$\pm$0.3 & 35 & 96   \\ \\
     \multirow{2}{*}{WESAD} & 0.2 & 4 & 94.5$\pm$0.4 & 9 & 50  \\
     & 0.3 & 3 & 67.4$\pm$0.4 & 7 & 63 \\
    \bottomrule
    \end{tabular}
    \label{tab:faulty}
\end{table}
\begin{table}[ht]
    \centering
    \caption{Local \ac{FS} from competitors approaches}
    \begin{tabular}{lll}
    \toprule
       Dataset  & Method & Local FS intersection \\
       & & (\%) \\
       \midrule
        \multirow{3}{*}{MAV (k=18)} & MRMR & 0 \\
         & ANOVA & 0 \\
         & HJMI & 0 \\
        \multirow{3}{*}{WESAD (k=4)} & MRMR & 100 \\
         & ANOVA & 100 \\
         & HJMI & 100 \\
    \bottomrule
    \end{tabular}
    \label{tab:local_fs_comp}
\end{table}
In Table \ref{tab:local_fs_comp} we show how the benchmark methods MRMR, ANOVA and HJMI behave when run in isolation on local dataset. Each method has been configured to select the optimal number of features found in a centralised setting. This is clearly an unrealistic situation that we use to demonstrate the limitations coming from running a non-\ac{FFS} algorithm in federated settings (i.e., on partial datasets). Precisely, taking into account the MAV dataset, all the algorithms run in isolation on each AV, find a different subset of features (i.e., null pairwise intersection). No agreement between AVs on the subset of features means that all the local data must be transmitted to the \ac{ES}, causing a non negligible waste of network resources. We motivate this behaviour with the fact that the original subset of features is redundant, as in MAV, running the \ac{FS} in isolation on portions of data is not a winning strategy. Conversely, when the original subset of features is less noisy, as in WESAD, it is more likely that all the AVs find, completely by chance, the same subset of features, i.e., without a way to coordinate the features selection in a consistent and provable way, there are no guarantees for the AVs to identify a consistent and shared subset of features.

Finally we analyse the \ac{FFS} performance in presence of \emph{faulty nodes}, i.e., a node experiencing issues in transmitting successfully its updates to the \ac{ES}. Note that, the causes preventing the updates' transmission might relate to either communication-related (i.e, a noisy channel) or the presence of a power-saving policy regulating the duty cycle of \acp{AV} switching off the network interface for a time corresponding to a communication round. The aim is assessing the robustness of \ac{FFS} when few nodes cannot contribute to the distributed learning at each communication round. To this end, we simulate a scenario where, at each communication round, a random number of \acp{AV} fail to communicate their updates to the \ac{ES}. We model the fault of a \ac{AV} performing a random draw from a Bernoulli distributed random variable, with parameter $\rho$. At the beginning of the simulation we set $\rho$ and, for each communication round and for each node, we perform a random draw, where 0 means faulty and 1 means non-faulty. This means that the updates of a faulty \ac{AV} are not considered for the execution of Algorithm \ref{algo:federated_algo_server}. We consider a fault rate $\rho$ equal to 0.2 and 0.3, meaning that at each round there are, on average, 2 and 3 faulty \acp{AV} out of 10, respectively. Such values can be reasonably assumed as upper bounds to evaluate the performance of the system. Higher rates would reveal that the scenario is not reasonably set up to run with any sort of reliability.

In Table~\ref{tab:faulty}  we report the performance of \ac{FFS}, for both datasets. For the MAV dataset, although \ac{FFS} selects $2\times$ and $3.3\times$ more features than the case when all the \acp{AV} contribute to the process (see Table\ref{tab:mav-cfsvsffs}), the compression rate deteriorates by 1\% and 3\%, respectively. The quality of the selection is confirmed by the accuracy that is statistically equivalent to the case without faulty \acp{AV}. Regarding WESAD, we notice that for $\rho=0.2$ \ac{FFS} performance is equivalent to the case with all non-faulty \acp{AV}. Conversely, for $\rho=0.3$ \ac{FFS} selects a smaller (i.e., 3 instead of 4) and less informative subset of features, as confirmed by the accuracy degradation. The reason is that the information collected by \ac{ES} at each round is not enough to select, globally, the most informative features. A final comment is about the network overhead, which can be further reduced, limiting the number of \emph{contributing} \acp{AV} during a communication round. In fact, Table~\ref{tab:faulty} suggests that there is a trade-off between accuracy, compression rate, and number of \emph{contributing} \acp{AV} through which we might optimise both the compression and the resources spent to find it. Moreover, there is a limit below which saving resources becomes detrimental to the learning process. However, understanding the nature of such a trade-off is left to future works.

\section{Conclusion and future directions}
\label{sec:conclusion}
The increasing development of \acp{ADS} can leverage AI to abstract both services and applications from the details of fast-flowing low-level data, such as sensor feeds. According to the Edge computing paradigm, a cyber physical system, namely \ac{AV}, is deputed in collecting data from sensors and perform a lightweight round of computation, by extracting features from raw data and selecting those that maximise the knowledge on the learning task.
Since the data gathering process is performed locally by each \ac{AV}, the selected features might represent a partial subset of those that characterize the phenomenon and might be inconsistent to learn the model of the underlying process.
We tackle this problem, by proposing a novel Federated Feature Selection (FFS) algorithm, exploiting a distributed computing paradigm applied to  \acp{AV}. In \ac{FFS}, \acp{AV} collaborate to iteratively come up with the minimal set of features selected from their local datasets, to be used as a compression schema for transmitting their data to the Edge Server. Feature selection is done by leveraging on the Mutual Information metric and the solution of the optimization problem is achieved through Cross-entropy method. The aggregation algorithm of the \ac{FFS} solution is based on a Bayesian approach through which we merge the control information sent by the \acp{AV} to the \ac{ES}. To test the proposed \ac{FFS} algorithm we presented two different learning tasks, by using real-world datasets: MAV and WESAD. 
The former was suitable to test \ac{FFS} with images and inertial
measurements, which characterize the position of an AV in the environment. The latter was suitable to characterize time series produced by human state monitoring systems, like ECG, EDA, EMG, etc. The results show that our \ac{FFS} algorithm identifies a minimal subset of informative features without sharing any raw data between AVs in the process. \ac{FFS} 
is robust to feature redundancy, i.e., in presence of high rates of redundant features, all the \acp{AV} can reach a consensus on the \ac{FS} achieving a compression rate up to 90x on the selected datasets. Finally, the quality of the feature selection is maintained, i.e., a learning model trained on the selected features is as accurate as a model trained on the whole feature set. Concluding, the proposed framework is general and modular, i.e., it can be applied to every incremental \ac{FS} algorithm that associates a probability to each feature. We plan to investigate how to turn it into a framework to include more \ac{FS} algorithms. Moreover, our solution is built on few simplifying assumptions: local datasets are \textit{iid} and data are labelled. Therefore, for the future we plan to extend it to include non-\textit{iid} data in possibly unsupervised or semi-supervised scenarios.   


\section*{Acknowledgment}
This work is partially supported by the MIUR PON project OK-INSAID (GA \#ARS01\_00917) and by the H2020 projects TEACHING (GA \#871385), HumanAI-Net, (GA \#952026), MARVEL (GA \#957337), and SoBigData++ (GA \#871042). 

\appendices 
\section*{Appendices}
\subsection{Proof of convergence of the federated method}
\label{appendixa}
In this section, we analyze the probability that the distribution $\mathbf{p}$ converges toward the optimal solution $\mathbf{p}^*$, when the Algorithm \ref{algo:main_CE} is applied in a centralized way. Then, we extend this result for the proposed federated algorithm.

The convergence analysis is based on the results in \cite{Costa2007,Wu2014}: following that notation, we introduce some preliminary definitions.
In the \ac{CE}, the candidate solutions $\mathbf{Z}_t=\{\mathbf{z}_1 \cdots \mathbf{z}_S\}$ generated at iteration $t$ are \textit{iid} with distribution $g(\mathbf{z},\mathbf{p}_{t-1})$.

We define $\mathcal{Z}_t:=\{\mathbf{z}_{j,\tau} \neq \mathbf{z^*}\; j=1 \cdots S,\;\tau=1 \cdots t\} \subseteq  \mathbf{Z}_t$  as the subset of $\mathbf{Z}_t$ of the samples generated up to $t$ that do not provide the optimal solution $\mathbf{z^*}$. The probability $\mathrm{Pr}(\mathcal{Z}_{t})$ that the optimal solution is not available until $t$ can be found as in the following:
\begin{gather}
    \nonumber
    \mathrm{Pr}(\mathcal{Z}_{t})=\mathrm{Pr}(\mathcal{Z}_1)\displaystyle\prod_{\tau=2}^{t} \mathrm{Pr}(\mathcal{Z}_{\tau}|\mathcal{Z}_{\tau-1})=\\
    \label{eq:pr_not_opt_sol}
    \mathrm{Pr}(\mathcal{Z}_1)\displaystyle\prod_{\tau=2}^{t} \left(\mathrm{Pr}(\mathbf{z}_{\tau}\neq\mathbf{z}^*|\mathcal{Z}_{\tau-1})\right)^S
\end{gather}

The equation (\ref{eq:pr_not_opt_sol}) comes from the statistical independence of $S$ identically distributed samples generated by the algorithm at iteration $t$. The upper bound for the probability $ \mathrm{Pr}(\mathbf{z}_{\tau}\neq\mathbf{z}^*|\mathcal{Z}_{\tau-1})$ that the optimal solution was unavailable until $\tau$ is derived in \cite{Costa2007,Wu2014} as:
\begin{equation}
\mathrm{Pr}(\mathbf{z}_{\tau}\neq\mathbf{z}^*|\mathcal{Z}_{\tau-1})\leq 1-\mathrm{Pr}(\mathbf{z}_1=\mathbf{z}^*)\displaystyle\prod_{i=1}^{\tau-1} (1-\alpha_i)^m
\label{eq:bu}
\end{equation}

where 
\begin{multline}
\mathrm{Pr}(\mathbf{z}_1=\mathbf{z}^*)=\\ \displaystyle\prod_{i=1}^m \bigg(p_i(z_i=0)\mathcal{I}{(z_i^*=1)}+\big(1-p_i(z_i=0)\big)\mathcal{I}{(z_i^*=0})\bigg)
\label{eq:tu}
\end{multline}
Note that due to the definition of $\mathcal{Z}_1$ its probability is $\mathrm{Pr}(\mathcal{Z}_1)=1-\mathrm{Pr}(\mathbf{z}_1=\mathbf{z}^*)$.

Combining equations \eqref{eq:bu} and \eqref{eq:tu}, equation \eqref{eq:pr_not_opt_sol} becomes:
\begin{multline}
\mathrm{Pr}(\mathbf{z}_{t} \neq \mathbf{z^*})\leq \\
\bigg(1-\displaystyle\prod_{i=1}^m \big(p_i(z_i=0)\mathcal{I}(z_i^*=1)+\big(1-p_i(z_i=0)\big)\mathcal{I}(z_i^*=0)\big)\bigg)\cdot\\
\displaystyle\prod_{\tau=2}^{t} \bigg( 1-\displaystyle\prod_{i=1}^m \big(p_i(z_i=0)\mathcal{I}(z_i^*=1)+\\+\big(1-p_i(z_i=0)\big)\mathcal{I}(z_i^*=0)\big)\displaystyle\prod_{j=1}^{\tau-1}
 (1-\alpha_j)^m\bigg)^S
\label{eq:pr_not_opt_sol2}
\end{multline}

The right side of the equation (\ref{eq:pr_not_opt_sol2}) is close to $0$ for $ t\rightarrow \infty $, if $\displaystyle\sum_{\tau=1}^{\infty}\displaystyle\prod_{j=1}^{\tau-1}(1-\alpha_j)^m \rightarrow \infty$, i.e., the sequence of the parameters $\alpha_j$ are generated by the function $\frac{1}{j\cdot m}$, as proven by authors in \cite{Knopp1956} (section 3.7). Note that, equation (\ref{eq:pr_not_opt_sol2}) can be used to
determine numerically a combination of parameter values that yields a desired minimum
probability of generating the optimal solution within a time $t$.

Therefore, Algorithm (\ref{algo:main_CE}) definitely provides the optimal solution when applied in a centralized way. We extend this result for the federated approach as follows. The \acp{AV} draw distinct samples $\mathbf{z}_1\cdots \mathbf{z}_S$ independently from an identical distribution, as stated in the section \ref{sec:fed_approach}. This means that the node $l$ finds an optimal solution for its $\mathbf{z}_1\cdots \mathbf{z}_S$ that differs for that obtained by the centralized algorithm. Hence, combining the local distributions into the global one as in equation (\ref{eq:joint_probability}), we need to prove that the local node can receive from the server a federated solution that is close to the solution provided by  the centralized scenario, for $t\rightarrow \infty$.

Defining the Hamming's distance $\mathcal{L}(\mathbf{z}^*,\mathbf{z}_{\tau})$ between the sample $\mathbf{z}_{\tau}$ at the time $\tau$ and the optimal solution $\mathbf{z}^*$, the set  $\tilde{\mathcal{Z}}_t:=\{\mathbf{z}_{i,\tau}\:|\:\mathcal{L}(\mathbf{z}^*,\mathbf{z}_{i,\tau})=m_l\}$ contains the samples generated up to time $t$ that differs for $m_l$ entries  from the optimal solution $\mathbf{z}^*$. 

As in (\ref{eq:pr_not_opt_sol}), we can calculate the probability $\mathrm{Pr}(\tilde{\mathcal{Z}}_t)$ as follows:

\begin{equation}
\mathrm{Pr}(\tilde{\mathcal{Z}}_t)=\mathrm{Pr}(\tilde{\mathcal{Z}}_1)\displaystyle\prod_{\tau=2}^{t} \mathrm{Pr}(\tilde{\mathcal{Z}}_{\tau}|\tilde{\mathcal{Z}}_{\tau-1})\\
\label{eq:pr_not_opt_sol_federated}   
\end{equation}
 
 Exploiting again the results in  \cite{Costa2007,Wu2014}, and the statistical independence of the $S$ identically distributed samples generated by the algorithm at a given iteration, the following equation holds for the conditional probability for the given node $l$:

\begin{multline}
\label{eq:prob_not_opt_sol_cond}
\mathrm{Pr}(\tilde{\mathcal{Z}}_{\tau}|\tilde{\mathcal{Z}}_{\tau-1})=\Big[\Big.\binom{m}{m_l}\mathrm{Pr}(\mathbf{z}_1=\mathbf{z}_1^*)\cdot\\
\displaystyle\prod_{i=1}^{\tau-1}(1-\alpha_{i,l})^{m-m_l}\cdot
\big(1-\mathrm{Pr}(\mathbf{z}_1=\mathbf{z}_1^*)\displaystyle \prod_{i=1}^{\tau-1}(1-\alpha_{i,l})^{m_l}\big)\Big.\Big]^S\\
    \end{multline}


Note that, the result provided in equation (\ref{eq:pr_not_opt_sol_federated}) refers to the $l$-th node. Hence, the global solution is obtained as the weighted average over all the local probabilities $\mathrm{Pr}(\tilde{\mathcal{Z}}_{l,t})$ as:  
 
\begin{equation}
    \mathrm{Pr}_{\mathbf{G}}(\tilde{\mathcal{Z}}_{t})=\displaystyle \sum_{l=1}^L \mathrm{Pr}(\tilde{\mathcal{Z}}_{l,t})\omega_l
\label{eq:pr_not_opt_sol_global}
\end{equation}

where $\omega_l$ are computed as in equation \eqref{eq:probability_data_from_node}.

The probability in (\ref{eq:pr_not_opt_sol_global}) is close to $0$, for $t\rightarrow \infty $, if $\displaystyle\sum_{\tau=1}^{\infty}\displaystyle\prod_{i=1}^{\tau-1}(1-\alpha_{i,l})^{m_l} \rightarrow \infty \; \forall \;l=1,\ldots,L$. Note that, if the sum of products of $(1-\alpha_{i,l})^{m_l}$ is close to $\infty$ also the sum of products of $(1-\alpha_{i,l})^{m-m_l}$ is close to $\infty$. The sequences of the $\alpha_{i,l}$ parameters guarantee the convergence also in this case. Indeed, the parameters are generated locally by the node, using the function $\frac{1}{m\cdot t}$.
\subsection{Analysis of the global probability computational effort}
\label{appendixb}
In this section, we analyze the probability distribution of the number of iterations $t$ needed to evaluate the global probability in (\ref{eq:joint_probability}). We address this issue by exploiting the result in (\ref{eq:pr_not_opt_sol_global}), which describes the probability that the global solution obtained at the iteration $t$ differs by $m_l$ entries from the optimal one. Hence, the probability that the global solution is reached within $t$ can be written as follows:

\begin{gather}
    \mathrm{Pr}_{\mathbf{G}}(\mathbf{z}_{t}= \mathbf{z^*})=1-\displaystyle \sum_{m_l=1}^m  \mathrm{Pr}_{\mathbf{G}}(\tilde{\mathcal{Z}}_{t})
\label{eq:pr_opt_sol_global_at_t}
\end{gather}

We can exploit the following inequality $(1-\alpha)^m \leq e^{-\alpha m}\; |\; 0\leq \alpha \leq 1,\; m\geq0$ to find  the upper bound shown in the following:

\begin{gather}
\nonumber
    \mathrm{Pr}_{\mathbf{G}}(\mathbf{z}_{t}=\mathbf{z^*}) \leq\\
\nonumber
1-\displaystyle \sum_{m_l=1}^m \sum_{l=1}^L \mathrm{Pr}(\tilde{\mathcal{Z}}_1)\displaystyle\prod_{\tau=2}^{t} \left[\binom{m}{m_l}\mathrm{Pr}(z_1=z_1^*)\right.\\
\nonumber
\left(exp\left(-\displaystyle\sum_{i=1}^{\tau-1}\alpha_{i,l}(m-m_l)\right)-\right. \\ 
\left.\left. \mathrm{Pr}(z_1=z_1^*)\:exp\left(-\displaystyle \sum_{i=1}^{\tau-1}\alpha_{i,l}\:m\right)\right)\right]^S \omega(l)
\label{eq:lower_pr_opt_sol_global_at_t}
\end{gather}

The difference between exponentials in equation (\ref{eq:lower_pr_opt_sol_global_at_t}) goes to zero faster than the binomial coefficient goes to infinity, as $m$ increases, if the coefficient $\alpha$ satisfies the conditions verified in the previous appendix.  Thus equation (\ref{eq:lower_pr_opt_sol_global_at_t}) can be used to evaluate the probability distribution of the number of iterations $t=1,2\ldots,\infty$ required to converge to the optimal global solution.
The numerical analysis shows an average value of $13$ iterations to converge by using the parameters presented in the section \ref{sec:num_results}, which is affordable for many edge devices like Nvidia Jetson Nano or RaspberryPi.     


%




\ifCLASSOPTIONcaptionsoff
  \newpage
\fi



\IEEEtriggeratref{0}
\bibliographystyle{IEEEtran}
\bibliography{biblio}

\end{document}